\documentclass{article}
\usepackage{footmisc}

\usepackage[preprint]{nips_2018}

\usepackage[utf8]{inputenc} 
\usepackage[T1]{fontenc}    

\PassOptionsToPackage{hyphens}{url}
\usepackage{multirow}
\usepackage{hyperref}
\usepackage{booktabs}       
\usepackage{amsfonts}       
\usepackage{nicefrac}       
\usepackage{microtype}      
\usepackage{xcolor}
\usepackage{graphicx}
\usepackage{listings}

\usepackage{float}
\usepackage{color}
\usepackage{mathcomp}
\usepackage{caption}
\usepackage[font=small,labelfont=bf]{caption}

\usepackage{amsmath}

\usepackage{multirow}
\usepackage{booktabs}

\title{Snips Voice Platform: an embedded \\ Spoken Language Understanding system \\for private-by-design voice interfaces}

\author{
\begin{tabular}{ccc}
~\\
Alice Coucke & Alaa Saade & Adrien Ball \\
~\\
Th\'eodore Bluche & Alexandre Caulier & David Leroy \\
~\\
Cl\'ement Doumouro & Thibault Gisselbrecht & Francesco Caltagirone \\
~\\ 
Thibaut Lavril & Ma\"{e}l Primet & Joseph Dureau \\
~\\
& {\normalfont Snips} & \\
& {\normalfont Paris, France}
\end{tabular}
}

\begin{document}

\maketitle

\definecolor{darkspringgreen}{rgb}{0.09, 0.45, 0.27}
\definecolor{cadetblue}{rgb}{0.36, 0.54, 0.66}
\definecolor{pureblue}{rgb}{0.0, 0.0, 1.0}
\definecolor{pastelgreen}{rgb}{0.47, 0.87, 0.47}
\definecolor{blush}{rgb}{0.87, 0.36, 0.51}
\definecolor{darkpastelpurple}{rgb}{0.59, 0.44, 0.84}
\definecolor{richelectricblue}{rgb}{0.03, 0.57, 0.82}
\definecolor{mikadoyellow}{rgb}{1.0, 0.77, 0.05}
\definecolor{neonfuchsia}{rgb}{1.0, 0.25, 0.39}
\definecolor{lightseagreen}{rgb}{0.13, 0.7, 0.67}
\definecolor{darkgreen}{rgb}{0.09, 0.51, 0.07}
\definecolor{cadmiumorange}{rgb}{0.93, 0.53, 0.18}
\newcommand{\red}[1]{\textcolor{blush}{#1}}
\newcommand{\blue}[1]{\textcolor{cadetblue}{#1}}
\newcommand{\bluetwo}[1]{\textcolor{richelectricblue}{#1}}
\newcommand{\pureblue}[1]{\textcolor{pureblue}{#1}}
\newcommand{\purple}[1]{\textcolor{darkpastelpurple}{#1}}
\newcommand{\green}[1]{\textcolor{pastelgreen}{#1}}
\newcommand{\greentwo}[1]{\textcolor{lightseagreen}{#1}}
\newcommand{\darkgreen}[1]{\textcolor{darkgreen}{#1}}
\newcommand{\yellow}[1]{\textcolor{mikadoyellow}{#1}}
\newcommand{\fuchsia}[1]{\textcolor{neonfuchsia}{#1}}
\newcommand{\orange}[1]{\textcolor{cadmiumorange}{#1}}

\begin{abstract}
  This paper presents the machine learning architecture of the Snips Voice Platform, a software solution to perform Spoken Language Understanding on microprocessors typical of IoT devices. The embedded inference is fast and accurate while enforcing privacy by design, as no personal user data is ever~collected. Focusing on Automatic Speech Recognition and Natural Language Understanding, we detail our approach to training high-performance Machine Learning models that are small enough to run in real-time on small devices. Additionally, we describe a data generation procedure that provides sufficient, high-quality training data without compromising user privacy.
\end{abstract}

\section{Introduction}
\label{intro}

Over the last years, thanks in part to steady improvements brought by deep learning approaches to speech recognition~\cite{mohamed2012acoustic,hinton2012deep,graves2013speech,bahdanau2016end}, voice interfaces have greatly evolved from spotting limited and predetermined keywords to understanding arbitrary formulations of a given intention. They also became much more reliable, with state-of-the-art speech recognition engines reaching human level in English~\cite{xiong2016achieving}. This achievement unlocked many practical applications of voice assistants which are now used in many fields from customer support~\cite{customersupport1,customersupport2}, to autonomous cars~\cite{autonomouscars}, or smart homes~\cite{smarthome1, smarthome2}. In particular, smart speaker adoption by the public is on the rise, with a recent study showing that nearly 20\% of U.S. adults reported having a smart speaker at home\footnote{\url{https://www.voicebot.ai/2018/03/07/new-voicebot-report-says-nearly-20-u-s-adults-smart-speakers/}}.

These recent developments however raise questions about user privacy -- especially since unique speaker identification is an active field of research using voice as a sensitive biometric feature~\cite{voicebiometrics}. The CNIL (French Data Protection Authority)\footnote{In French: \url{https://www.cnil.fr/fr/enceintes-intelligentes-des-assistants-vocaux-connectes-votre-vie-privee}} advises owners of connected speakers to switch off the microphone when possible and to warn guests of the presence of such a device in their home. The General Data Protection Regulation which harmonizes data privacy laws across the European Union\footnote{\url{https://www.eugdpr.org/}} indeed requires companies to ask for explicit consent before collecting user data.

Some of the most popular commercial solutions for voice assistants include Microsoft's \textit{Cortana}, Google's \textit{DialogFlow}, IBM's \textit{Watson}, or Amazon \textit{Alexa}~\cite{ASK17}. In this paper, we introduce a competing solution, the Snips Voice Platform which, unlike the previous ones, is completely cloud independent and runs offline on typical IoT microprocessors, thus guaranteeing privacy by design, with no user data ever collected nor stored. The Natural Language Understanding component of the platform is already open source \cite{SnipsNLU}, while the other components will be opensourced in the future.

The aim of this paper is to contribute to the collective effort towards ever more private and efficient cloud-independent voice interfaces. To this end, we devote this introduction to a brief description of the Snips Ecosystem and of some of the design principles behind the Snips Voice Platform.

\subsection{The Snips Ecosystem}
\label{sec:snips_ecosystem}
The Snips ecosystem comprises a web console\footnote{\url{https://console.snips.ai}} to build voice assistants and train the corresponding Spoken Language Understanding (SLU) engine, made of an Automatic Speech Recognition (ASR) engine and a Natural Language Understanding (NLU) engine. The console can be used as a self-service development environment by businesses or individuals, or through professional services. The Snips Voice Platform is free for non-commercial use. Since its launch in Summer 2017, over 23,000 Snips voice assistants have been created by over 13,000 developers. The languages currently supported by the Snips platform are English, French and German, with additional NLU support for Spanish and Korean. More languages are added regularly.

An \textit{assistant} is composed of a set of \textit{skills} -- e.g. \texttt{SmartLights}, \texttt{SmartThermostat}, or \texttt{SmartOven} skills for a \texttt{SmartHome} assistant -- that may be either selected from preexisting ones in a \textit{skill store} or created from scratch on the web console. A given skill may contain several \textit{intents}, or user intention -- e.g. \texttt{SwitchLightOn} and \texttt{SwitchLightOff} for a \texttt{SmartLights} skill. Finally, a given intent is bound to a list of \textit{entities} that must be extracted from the user's query -- e.g. \texttt{room} for the \texttt{SwitchLightOn} intent. We call \emph{slot} the particular value of an entity in a query -- e.g. \texttt{kitchen} for the entity \texttt{room}. When a user speaks to the assistant, the SLU engine trained on the different skills will handle the request by successively converting speech into text, classifying the user's intent, and extracting the relevant slots.

\begin{figure}[ht]
\vskip 0.2in
\begin{center}
\centerline{\includegraphics[width=\columnwidth]{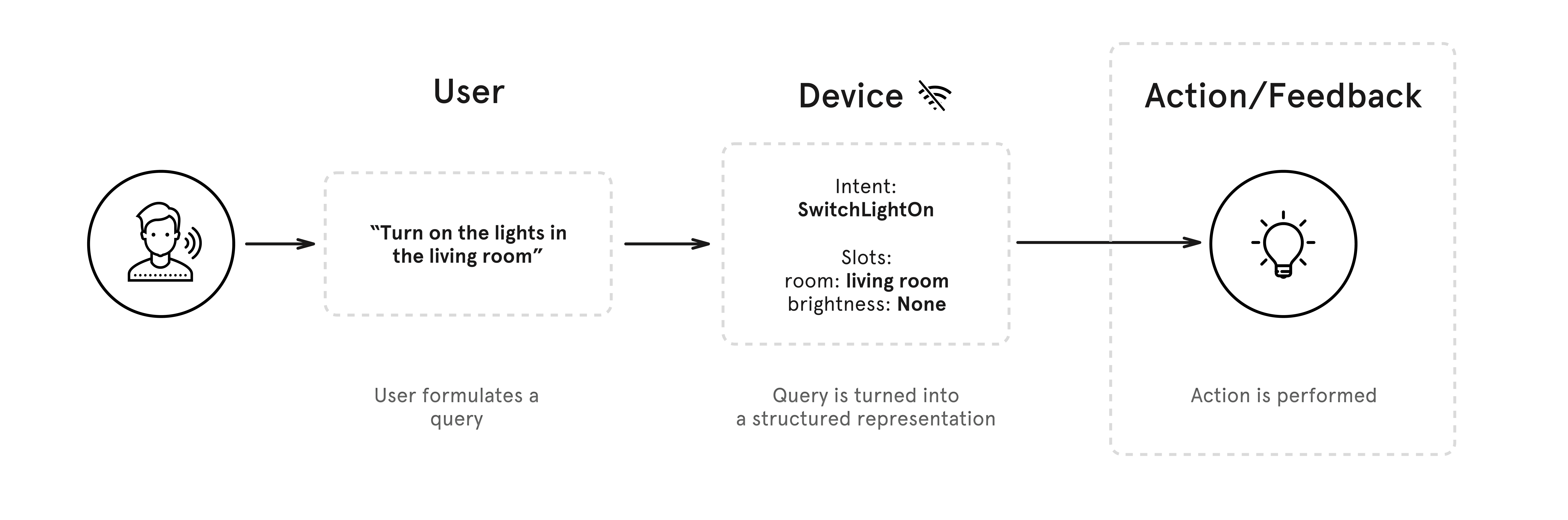}}
\caption{Interaction flow}
\label{fig:flow}
\end{center}
\vskip -0.2in
\end{figure}

Once the user's request has been processed and based on the information that has been extracted from the query and fed to the device, a dialog management component is responsible for providing a feedback to the user, or performing an \textit{action}. It may take multiple forms, such as an audio response via speech synthesis or a direct action on a connected device~-- e.g. actually turning on the lights for a \texttt{SmartLights} skill. Figure~\ref{fig:flow} illustrates the typical interaction flow.

\subsection{A private-by-design embedded platform}

The \emph{Privacy by Design} \cite{cavoukian2009privacy,langheinrich2001privacy} principle sets privacy as the default standard in the design and engineering of a system. In the context of voice assistants, that can be deployed anywhere including users' homes, this principle calls for a strong interpretation to protect users against any future misuse of their private data. In the following, we call \emph{private-by-design} a system that does not transfer user data to any remote location, such as cloud servers.

Within the Snips ecosystem, the SLU components are trained on servers, but the inference happens directly on the device once the assistant has been deployed - no data from the user is ever collected nor stored. This design choice adds engineering complexity as most IoT devices run on specific hardware with limited memory and computing power. Cross-platform support is also a requirement in the IoT industry, since IoT devices are powered by many different hardware boards, with sustained innovation in that field.

For these reasons, the Snips Voice Platform has been built with portability and footprint in mind. Its embedded inference runs on common IoT hardware as light as the Raspberry Pi 3 (CPU with 1.4 GHz and 1GB of RAM), a popular choice among developers and therefore our reference hardware setting throughout this paper. Other Linux boards are also supported, such as IMX.7D, i.MX8M, DragonBoard 410c, and Jetson TX2. The Snips SDK for Android works with devices with Android 5 and ARM CPU, while the iOS SDK targets iOS 11 and newer. For efficiency and portability reasons, the algorithms have been re-implemented whenever needed in Rust~\cite{Rust14} -- a modern programming language offering high performance, low memory overhead, and cross-compilation.

\begin{figure}[ht]
\begin{center}
\centerline{\includegraphics[width=\columnwidth]{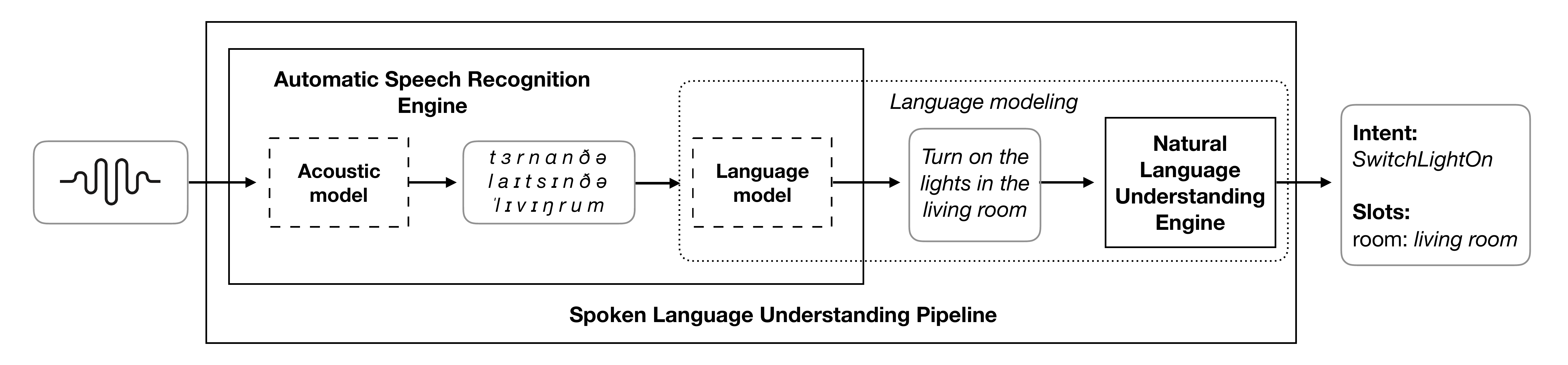}}
\caption{Spoken Language Understanding pipeline}
\label{fig:ml_flow}
\end{center}
\end{figure}

SLU engines are usually broken down into two parts: Automatic Speech Recognition (ASR) and Natural Language Understanding (NLU). The ASR engine translates a spoken utterance into text through an acoustic model, mapping raw audio to a phonetic representation, and a Language Model (LM), mapping this phonetic representation to text. The NLU then extracts intent and slots from the decoded query. As discussed in section~\ref{sec:lmu}, LM and NLU have to be mutually consistent in order to optimize the accuracy of the SLU engine. It is therefore useful to introduce a \emph{language modeling} component composed of the LM and NLU.
Figure~\ref{fig:ml_flow} describes the building blocks of the SLU pipeline.

As stated above, ASR engines relying on large deep learning models have improved drastically over the past few years. Yet, they still have a major drawback today. For example, the model achieving human
parity in~\cite{xiong2016achieving} is a combination of several neural networks,
each containing several hundreds of millions of parameters, and large-vocabulary language models made of several millions of n-grams. The size of these models, along with the computational resources necessary to run them in real-time, make them unfit for deployment on small devices, so that solutions implementing them are bound to rely on the cloud for speech recognition.

 Enforcing privacy by design therefore implies developing new tools to build reliable SLU engines that are constrained in size and computational requirements, which we detail in this paper. In section~\ref{sec:acoustic}, we describe strategies to obtain small (10MB) and robust acoustic models trained on general speech corpora of a few hundred to a few thousand hours. Section~\ref{sec:lmu} is devoted to a description of the language modeling approach of the Snips SLU engine. Notably, we show how to ensure consistency between the language model of the ASR engine and the NLU engine while specializing them to a particular use case. The resulting SLU engine is lightweight and fast to execute, making it fit for deployment on small devices and the NLU component is open source \cite{SnipsNLU}. In section~\ref{sec:end-to-end-evaluation}, we illustrate the high generalization accuracy of the SLU engine in the context of real-word voice assistants. Finally, we discuss in section~\ref{fig:datagen} a data generation procedure to automatically create training sets replacing user data.

\section{Acoustic model}
\label{sec:acoustic}

\newcommand\todotb[1]{\textbf{\color{red} TODO (theo) #1}}
\newcommand\wer{(Word Error Rate, \%)}
\newcommand\citend[1]{\textbf{\color{blue} [#1]}}
\newcommand\tab[1]{Table~\ref{tab:#1}}
\newcommand\fig[1]{Figure~\ref{fig:#1}}

The acoustic model is the first step of the SLU pipeline, and is therefore crucial to its functioning. If the decoding contains errors, it might compromise the subsequent steps and trigger a different action than that intended by the user.

The acoustic model is responsible for converting raw audio data to what can approximately be interpreted as phone probabilities, i.e. context-dependent clustered Hidden Markov Model (HMM) state probabilities. These probabilities are then fed to a language model, which decodes a sequence of words corresponding to the user utterance. The acoustic and language models are thus closely related in the Automatic Speech Recognition (ASR) engine, but are often designed and trained separately. The construction of the language model used in the SLU engine is detailed in section~\ref{sec:lmu}.

In this section, we present the acoustic model. First, we give details about
how the training data is collected, processed, cleaned, and augmented. Then, we present the acoustic model itself (a hybrid of Neural Networks and Hidden Markov Models, or NN/HMM)
and how it is trained. Finally, we present the performance of the acoustic model
in a large-vocabulary setup, in terms of word error rate (WER), speed, and memory usage.

\subsection{Data}

\paragraph*{Training data.}

To train the acoustic model, we need several hundreds to thousands of hours of audio data with corresponding transcripts. The data is collected from public or commercial sources. A realignment of transcripts to the audio is performed to match transcripts to timestamps. This additionally helps in removing transcription errors that might be present in the data.
The result is a set of audio extracts and matching transcripts, with lengths suitable for acoustic training (up to a few dozen seconds). This data is split in a training, testing, and development sets.

\paragraph*{Data augmentation.}

One of the main issues regarding the training of the acoustic model is the lack of data corresponding to real usage scenari. Most of the available training data is clear close-field speech, but voice assistants will often be used in noisy conditions (music, television, environment noise), from a distance of several meters in far-field conditions, and in rooms or cars with reverberation.
From a machine learning perspective, data corresponding to real usage of the system –- or in-domain data -- is extremely valuable. Since spoken utterances from the user are not collected by our platform for privacy reasons, noisy and reverberant conditions are simulated by augmenting the data. Thousands of virtual rooms of different sizes are thus generated with random microphone and speaker locations, and the rerecording of the original data in those conditions is simulated using a method close to that presented in~\cite{kim2017generation}.

\subsection{Model training}

Acoustic models are hybrid NN/HMM models. More specifically, they are a custom
version of the \texttt{s5} training recipe of the Kaldi toolkit~\cite{povey2011kaldi}. 40 MFCC features are extracted from the audio signal with windows of size 25ms every 10ms.

Models with a variable number of layers and neurons can be trained, which will impact their accuracy and computational cost. We can thus train different model architectures depending on the target hardware and the desired application accuracy. In the following evaluation (section~\ref{sec:acoustic_model_evaluation}), we present performance results of a model targeted for the Raspberry Pi 3.

First, a speaker-adaptive Gaussian Mixture Model Hidden Markov Model (GMM-HMM) is trained on the speech corpus to obtain a context-dependent bootstrapping model with which we align the full dataset and extract lattices to prepare the neural network training.

\begin{figure}[ht]
  \begin{center}
    \includegraphics[width=.5\columnwidth]{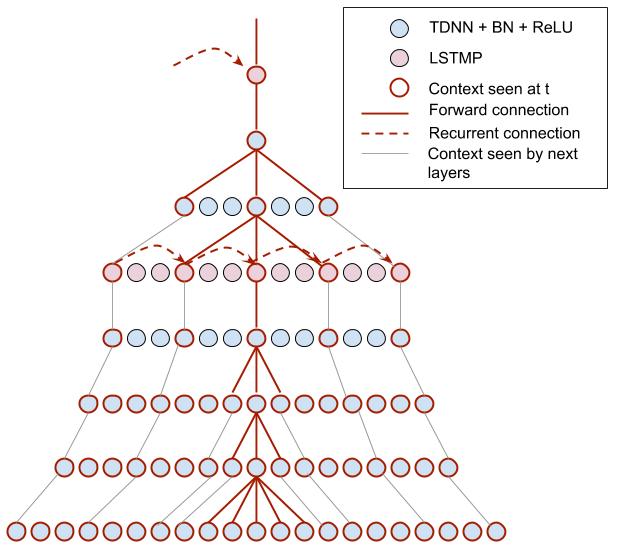}
  \end{center}
  \caption{The TDNN-LSTM architecture used in the presented models.\label{fig:nnetarchi}}
\end{figure}

\begin{table}[ht]
  \caption{Network architecture.
  The \textit{Context} denotes the number of relative frames seen by the layer at time t. For instance, the recurrent connections skip 3 frames in LSTMP layers. A projection layer size of $N$ is denoted \texttt{pN}.
  (TDNN: Time-Delay layer; LSTMP: Long Short-Term Memory with Projection layer).}
  \label{tab:nnetarchi}
  \begin{center}
      \begin{tabular}{rclll}
        \textbf{Layer Type} & \textbf{Context} & \texttt{nnet-256} & \texttt{nnet-512} & \texttt{nnet-768} \\\hline
        TDNN + BatchNorm + ReLU & \texttt{\{-2, -1, 0, 1, 2\}} & 256 & 512 & 768 \\
        TDNN + BatchNorm + ReLU & \texttt{\{-1, 0, 1\}} & 256 & 512 & 768 \\
        TDNN + BatchNorm + ReLU & \texttt{\{-1, 0, 1\}} & 256 & 512 & 768 \\
        LSTMP & rec:\texttt{-3} & 256, p128 & 512, p256 & 768, p256 \\
        TDNN + BatchNorm + ReLU & \texttt{\{-3, 0, 3\}} & 256 & 512 & 768 \\
        TDNN + BatchNorm + ReLU & \texttt{\{-3, 0, 3\}} & 256 & 512 & 768 \\
        LSTMP & rec:\texttt{-3} & 256, p128 & 512, p256 & 768, p256 \\\hline
        \textbf{Num. params} & & 2.6M & 8.7M & 15.4M \\\hline
      \end{tabular}
\end{center}
\end{table}

We train a deep neural network, consisting of time-delay layers similar to those presented in~\cite{tdnn}, and long short-term memory layers similar to those of~\cite{peddinti2018low}.
The architecture, close to that of~\cite{peddinti2018low}, is summarized
in~\fig{nnetarchi} and~\tab{nnetarchi}. The Raspberry Pi 3 model uses 7 layers, and is trained with the lattice-free Maximum Mutual Information (MMI) criterion~\cite{lfmmi},
using natural gradient descent, with a learning rate of 0.0005 and
the backstitching trick~\cite{backstitch}. We follow the approach described
in~\cite{lfmmi} to create fast acoustic models, namely an HMM topology with one state for each of the 1,700 context-dependent senones, operating at a third of the original frame rate.

\subsection{Acoustic model evaluation}
\label{sec:acoustic_model_evaluation}
In this section, we present an evaluation of our acoustic model for English. Our goal is the design of an end-to-end SLU pipeline which runs in real-time on small embedded devices, but has state-of-the-art accuracy. This requires tradeoffs between speed and the generality of SLU task. More precisely, we use domain-adapted language models described in section~\ref{sec:lmu}, to compensate for the decrease of accuracy of smaller acoustic models.

However in order to assess the quality of the acoustic model in a more general setting, the evaluation of this section is carried out in a large vocabulary setup, on the LibriSpeech evaluation dataset~\cite{panayotov2015Librispeech}, chosen because it is freely available and widely used in state-of-the-art comparisons. 
The language model for the large-vocabulary evaluation is also freely available online\footnote{\url{http://www.openslr.org/11/}}. It is a pruned trigram LM with a vocabulary of 200k words, trained on the content of public domain books.

Two sets of experiments are reported. In the first set, the models are trained
only on the LibriSpeech training set (or on a subset of it). It allows us to validate our
training approach and keep track of how the models we develop compare to the
state of the art when trained on public data. Then, the performance of the model in terms of speed and memory usage is studied, which allows us to select a good tradeoff for the targeted Raspberry Pi 3 setting.

\subsubsection{Model architecture trained and evaluated on LibriSpeech}

To evaluate the impact of the dataset and model sizes on the model accuracy, neural networks of different sizes are trained on different subsets of the
LibriSpeech dataset, with and without data augmentation. The results obtained
with \texttt{nnet-512} are reported in \tab{bignet-data}. The \textit{Num. hours}
column corresponds to the number of training hours (460h in the \textit{train-clean} split of the LibriSpeech dataset and
500h in the \textit{train-other} split). The data augmentation was only applied to the clean
data. For example \textit{460x2} means 460h of clean data + 460h of augmented
data.

\begin{table}[htb]
  \caption{Decoding accuracy of \texttt{nnet-512} with different amounts of training data \wer}
  \label{tab:bignet-data}
  \begin{center}
    \begin{tabular}{rcccc}
      \textbf{Num. hours} &   \textbf{dev-clean} & \textbf{dev-other} & \textbf{test-clean} & \textbf{test-other} \\
      \hline
      460       & 6.3 & 21.8 & 6.6 & 23.1 \\
      460x2     & 6.2 & 19.5 & 6.5 & 19.7 \\
      960       & 6.2 & 16.4 & 6.4 & 16.5 \\
      460x6+500 & 6.1 & 16.3 & 6.4 & 16.5 \\\hline
      KALDI     & 4.3 & 11.2 & 4.8 & 11.5 \\\hline
    \end{tabular}
  \end{center}
\end{table}

We observe that adding data does not have much impact on LibriSpeech's clean
test sets (dev-clean and test-clean). The WER however decreases when adding data on the datasets marked as \textit{other} (dev-other, test-other). In general (not shown in those tests), adding more data and using data augmentation increases significantly the performance on noisy and reverberant conditions.

In the next experiment, the neural networks are trained with the same architecture but different layer sizes on the 460x6+500 hours dataset. Results are reported in \tab{net-size}. This shows that larger models are capable of fitting the data and generalizing better, as expected. This allows us to choose the best tradeoff between precision and computational cost depending on each target hardware and assistant~needs.

\begin{table}[htb]
  \caption{Decoding accuracy of neural networks of different sizes \wer}
  \label{tab:net-size}
  \begin{center}
    \begin{tabular}{rcccc}
      \textbf{Model} & \textbf{dev-clean} & \textbf{dev-other} & \textbf{test-clean} & \textbf{test-other} \\
      \hline
      \texttt{nnet-256} & 7.3 & 19.2 & 7.6 & 19.6 \\
      \texttt{nnet-512} & 6.4 & 17.1 & 6.6 & 17.6 \\
      \texttt{nnet-768} & 6.4 & 16.8 & 6.6 & 17.5 \\\hline
              KALDI     & 4.3 & 11.2 & 4.8 & 11.5 \\\hline
    \end{tabular}
  \end{center}
\end{table}

\subsubsection{Online recognition performance}

While it is possible to get closer to the state of the art using larger neural network architectures, their associated memory and computational costs would prohibit their deployment on small devices. In section~\ref{sec:lmu}, we show how carefully adapting the LM allows to reach high end-to-end accuracies using the acoustic models described here. We now report experiments on the processing speed of these models on our target Raspberry Pi 3 hardware setting. We trained models with various sizes enjoying a faster-than-real-time processing factor, to account for additional processing time (necessitated e.g. by the LM decoding or the NLU engine), and chose a model with a good compromise of accuracy to real-time factor and model size (on disk and in RAM).

\begin{table}[htb]
  \caption{Comparison of speed and memory performance of
           \texttt{nnet-256} and \texttt{nnet-768}. RTF refers to real time ratio.}
  \label{tab:am-speed}
  \begin{center}
    \begin{tabular}{rccc}
      \textbf{Model} & \textbf{Num. Params (M)} & \textbf{Size (MB)} & \textbf{RTF (Raspberry Pi 3)}  \\
      \hline
      \texttt{nnet-256} &  2.6 & 10 & $ <1 $ \\
      \texttt{nnet-768} & 15.4 & 59 & $$ >1 $$ \\
    \end{tabular}
  \end{center}
\end{table}

As a reference, in terms of model size (as reported in \tab{am-speed}) \texttt{nnet-256} is nearly six times smaller than \texttt{nnet-768}, with 2.6M parameters vs 15.4M, representing 10MB vs 59MB on disk. The gain is similar in RAM. In terms of speed, the \texttt{nnet-256} is 6 to 10 times faster than the \texttt{nnet-768}. These tradeoffs and comparison with other trained models led us to select the \texttt{nnet-256}. It has a reasonable speed and memory footprint, and the loss in accuracy is compensated by the adapted LM and robust~NLU.

This network architecture and size will be the one used in the subsequent experiments. The different architecture variations presented in this section were chosen for the sake of comparison and demonstration. This experimental comparison, along with optional layer factorization (similar to~\cite{prabhavalkar2016compression}) or weight quantization are carried out for each target hardware setting, but this analysis is out of the scope of this paper.


\section{Language Modeling}
\label{sec:lmu}


We now turn to the description of the language modeling component of the Snips platform, which is responsible for the extraction of the intent and slots from the output of the acoustic model. This component is made up of two closely-interacting parts. The first is the language model (LM), that turns the predictions of the acoustic model into likely sentences, taking into account the probability of co-occurrence of words. The second is the Natural Language Understanding (NLU) model, that extracts intent and slots from the prediction of the Automatic Speech Recognition (ASR) engine.

In typical commercial large vocabulary speech recognition systems, the LM component is usually the largest in size, and can take up to terabytes of storage~\cite{chelba2012large}. Indeed, to account for the high variability of general spoken language, large vocabulary language models need to be trained on very large text corpora. The size of these models also has an impact on decoding performance: the search space of the ASR is expanded, making speech recognition harder and more computationally demanding. Additionally, the performance of an ASR engine on a given domain will strongly depend on the perplexity of its LM on queries from this domain, making the choice of the training text corpus critical. This question is sometimes addressed through massive use of users' private data~\cite{chelba2010query}.

One option to overcome these challenges is to \emph{specialize} the language model of the assistant to a certain domain, e.g. by restricting its vocabulary as well as the variety of the queries it should model. While this approach appears to restrict the range of queries that can be made to an assistant, we argue that it does not impair the usability of the resulting assistant. In fact, while the performance of an ASR engine alone can be measured using e.g. the word error rate as in the previous section, we assess the performance of the SLU system through its end-to-end, speech-to-meaning accuracy, i.e. its ability to correctly predict the intent and slots of a spoken utterance. As a consequence, it is sufficient for the LM to correctly model the sentences that are in the domain that the NLU supports. The size of the model is thus greatly reduced, and the decoding speed increases. The resulting ASR is particularly robust within the use case, with an accuracy unreachable under our hardware constraints for an all-purpose, general ASR model. In the following, we detail the implementation of this design principle, allowing the Snips SLU component to run efficiently on small devices with high accuracy, and illustrate its performance on two real-world assistants.

\subsection{Data}

In application of the principles outlined above, we use the same data to train both LM and NLU. The next section is devoted to a description of this dataset. The generation of this dataset is discussed in section~\ref{sec:datagen}.

\subsubsection{Training dataset}
\label{sec:asr-lm-training-set}

The dataset used to train both the LM and NLU contains written \emph{queries} exemplifying \emph{intents} that depend on \emph{entities}.

Entities are bound to an intent and used to describe all the possible values for a given attribute. For example, in the case of a \texttt{SmartLights} assistant handling connected lights, these entities are \texttt{room}, \texttt{brightness} and \texttt{color}. They are required by the assistant logic to execute the right action. Another example dealing with weather-related queries is described in section~\ref{sec:end-to-end-evaluation}. An intent often has several entities that it can share with other intents. For instance, the \texttt{room} entity is used by several intents (\texttt{SwitchLightOn} and \texttt{SwitchLightOff}), since the user might want to specify the room for both switching on and switching off the lights.

Entities can be of two types, either \emph{custom} or \emph{built-in}. Custom entities are user-defined entities that can be exhaustively specified by a list of values (e.g. \texttt{room}: kitchen, bedroom, etc. and \texttt{color}: blue, red, etc.). Built-in entities are common entities that cannot be easily listed exhaustively by a user, and are therefore provided by the platform (numbers, ordinals, amounts with unit, date and times, durations, etc.). In our \texttt{SmartLights} example, the entity \texttt{brightness} can be any number between 0 and 100, so that the built-in entity type \texttt{snips/number} can be used.

A query is the written expression of an intent. For instance, the query ``\texttt{set the kitchen lights intensity to 65}'' is associated with the intent \texttt{SetLightBrightness}. Slot labeling is done by specifying chunks of the query that should be bound to a given entity. Using the same examples, the slots associated with the \texttt{room} and \texttt{brightness} entities in the query can be specified as follows: ``\texttt{set the (kitchen)[room] lights intensity to (65)[brightness]}''. The number of queries per intent ranges from a few ones to several thousands depending on the variability needed to cover most common wordings.

\subsubsection{Normalization}

One key challenge related to end-to-end SLU is data consistency between training and inference. The dataset described above is collected via the console where no specific writing system, nor cleaning rules regarding non-alphanumeric characters are enforced. Before training the LM, this dataset therefore needs to be \emph{verbalized}: entity values and user queries are tokenized, normalized to a canonical form, and verbalized to match entries from a lexicon. For instance, numbers and dates are spelled out, so that their pronunciation can be generated from their written form. 

Importantly, we apply the same preprocessing before training the NLU. This step ensures consistency when it comes to inference. More precisely, it guarantees that the words output by the ASR match those seen by the NLU during training. The normalization pipeline is used to handle languages specificities, through the use of a class-based tokenizer that allows support for case-by-case verbalization for each token class. For instance, numeric values are transliterated to words, punctuation tokens skipped, while quantities with units such as amounts of money require a more advanced verbalization (in English, ``\texttt{\$25}'' should be verbalized as ``\texttt{twenty five dollars}''). The tokenizer is implemented as a character-level finite state transducer, and is designed to be easily extensible to accommodate new token types as more languages are supported.

\subsection{Language model}
\label{sec:asr-lm-language-model-building}

The mapping from the output of the acoustic model to likely word sequences is done via a Viterbi search in a weighted Finite State Transducer (wFST)~\cite{mohri2001weighted}, called \emph{ASR decoding graph} in the following. Formally, the decoding graph may be written as the composition of four wFSTs,

\begin{align}
H * C * L * G\, ,\label{eq:eager_hclg}
\end{align}

where $*$ denotes transducer composition (see section~\ref{sec:dynamic-lm}), $H$ represents Hidden Markov Models (HMMs) modeling context-dependent phones, $C$ represents the context-dependency, $L$ is the lexicon and $G$ is the LM, typically a bigram or a trigram model represented as a wFST. Determinization and minimization operations are also applied at each step in order to compute equivalent optimized transducers with less states, allowing the composition and the inference to run faster. More detailed definitions of the previous classical transducers are beyond the scope of this paper, and we refer the interested reader to~\cite{mohri2001weighted,povey2011kaldi} and references therein. In the following, we focus on the construction of the G transducer, encoding the LM, from the domain-specific dataset presented above.

\subsubsection{Language Model Adaptation}

As explained earlier, the ASR engine is required to understand arbitrary formulations of a finite set of intents described in the dataset. In particular, it should be able to generalize to unseen queries within the same domain, and allow entity values to be interchangeable. The generalization properties of the ASR engine are preserved by using a statistical n-gram LM~\cite{katz1987estimation} allowing to mix parts of the training queries to create new ones, and by using class-based language modeling~\cite{brown1992class} where the value of each entity may be replaced by any other. We now detail the resulting LM construction strategy.

\begin{figure}
  \begin{center}
    \includegraphics[width=1\columnwidth]{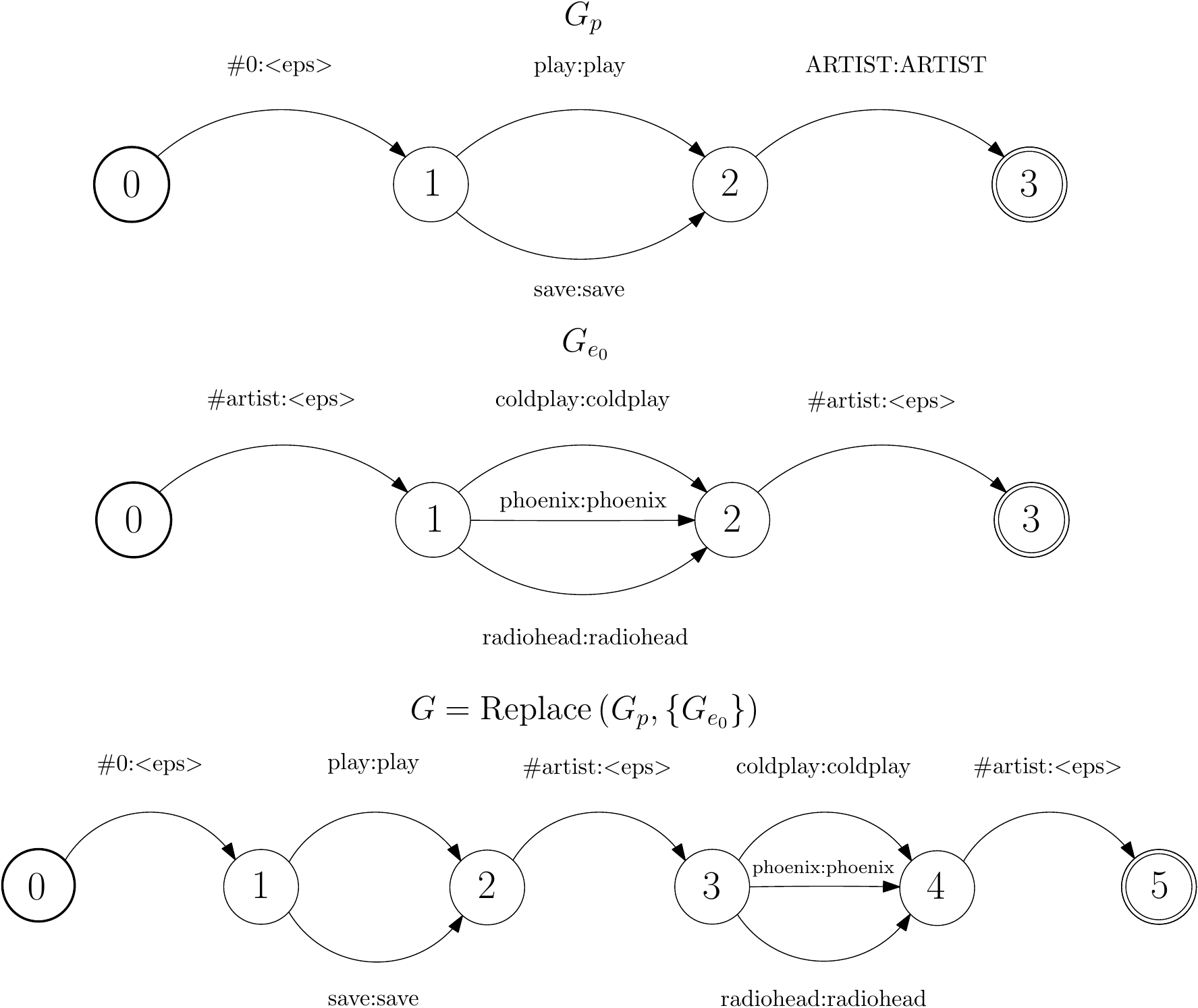}
  \end{center}{}
  \caption{Graphical representation of a class-based language model on a simple assistant understanding queries of the form ``\texttt{play ARTIST}''. The top wFST represents the pattern n-gram language model $G_p$. The middle wFST represent the ``\texttt{ARTIST}'' entity wFST called here $G_{e_0}$. The bottom wFST represents the result of the replacement operation~(\ref{eq:eager_replace}). Note that in order to keep the determinization of the HCLG tractable, a disambiguation symbol ``\texttt{\#artist}'' is added on the input and output arcs of $G_{e_0}$~\cite{horndasch2016add}. \label{fig:class-based}}
\end{figure}

The first step in building the LM is the creation of \emph{patterns} abstracting the type of queries the user may make to the assistant. Starting from the dataset described above, we replace all occurrences of each entity by a symbol for the entity. For example, the query ``\texttt{Play some music by (The Rolling Stones)[artist]}'' is abstracted to ``\texttt{Play some music by ARTIST}''. An n-gram model is then trained on the resulting set of patterns, which is then converted to a wFST called $G_{p}$~\cite{mohri2008speech}. Next, for each entity $e_i$ where $i\in[1, n]$ and $n$ is the number of entities, an acceptor $G_{e_i}$ is defined to encode the values the entity can take.
The construction of $G_{e_i}$ depends on the type of the entity. For custom entities, whose values are listed exhaustively in the dataset, $G_{e_i}$ can be defined either as a union of acceptors of the different values of the entity, or as an n-gram model trained specifically on the values of the entity. For built-in entities such as numbers or dates and times, $G_{e_i}$ is a wFST representation of a generative grammar describing the construction of any instance of the entity. The LM transducer $G$ is then defined as~\cite{horndasch2016add}

\begin{align}
G = \text{Replace}(G_{p}, \{G_{e_i}\, ,\forall i\in[1, n] \})\, ,\label{eq:eager_replace}
\end{align}

where $\text{Replace}$ denotes wFST replacement. For instance, in the example above, the arcs of $G_p$ carrying the ``\texttt{ARTIST}'' symbol are expanded into the wFST representing the ``\texttt{artist}'' entity. This process is represented on a simple LM on Figure~\ref{fig:class-based}.
The resulting $G$ allows the ASR to generalize to unseen queries and to swap entity values. Continuing with the simple example introduced above, the query ``\texttt{Play me some music by The Beatles}'' has the same weight as ``\texttt{Play me some music by The Rolling Stones}'' in the LM, while the sentence ``\texttt{Play music by The Rolling Stones}'' also has a finite weight thanks to the n-gram back-off mechanism.
The lexicon transducer $L$ encodes the pronunciations of all the words in both $G_p$ and $\{G_{e_i}\, ,\forall i\in[1, n] \}$. The pronunciations are obtained from large base dictionaries, with a fall-back to a statistical grapheme-to-phoneme (G2P) system~\cite{novak2016phonetisaurus} to generate the missing pronunciations.

\subsubsection{Dynamic Language Model}
\label{sec:dynamic-lm}
The standard way to compute the decoding graph is to perform compositions from right to left with the following formula
\begin{align}
\label{eq:hclg_static_ordered}
H * (C * (L * G))\, ,
\end{align}
where each composition is followed by a determinization and a minimization. The order in which the compositions are done is important, as the composition $(H * C * L) * G$ is known to be intractable when $G$ is not deterministic, as is the case when G is a wFST representing an n-gram model~\cite{mohri2001weighted}. We will refer to the result of equation~\ref{eq:hclg_static_ordered} as a \emph{static model} in the following.

In the context of embedded inference, a major drawback of this standard method is the necessity to compute and load the static HCLG decoding graph in memory in order to perform speech recognition. The size of this decoding graph can claim a large chunk of the $1$GB of RAM available on a Raspberry Pi 3, or even be too big for smaller devices. Additionally, since LMs are trained synchronously in the Snips web console after the user has created their dataset (see section~\ref{sec:snips_ecosystem}), it is important for the decoding graph to be generated as fast as possible.

For these reasons, a \emph{dynamic} language model, composing the various transducers upon request instead of ahead of time, is employed. This is achieved by replacing the compositions of equation~(\ref{eq:eager_hclg}) by delayed (or lazy) ones~\cite{allauzen2007openfst}. Consequently, the states and transitions of the complete HCLG decoding graph are not computed at the time of creation, but rather at runtime during the inference, notably speeding up the building of the decoding graph. Additionally, employing lazy composition allows to break the decoding graph into two pieces (HCL on one hand, and G on the other). The sum of the sizes of these pieces is typically several times smaller than the equivalent, statically-composed HCLG.

In order to preserve the decoding speed of the ASR engine using a dynamic language model, a better composition algorithm using lazy look-ahead operations must be used. Indeed, a naive lazy composition typically creates many non co-accessible states in the resulting wFST, wasting both time and memory. In the case of a static decoding graph, these states are removed through a final optimization step of the HCLG that cannot be applied in the dynamic case because the full HCLG is never built. This issue can be addressed through the use of composition filters~\cite{allauzen2009generalized,allauzen2010filters}. In particular, the use of look-ahead filters followed by label-reachability filters with weights and labels pushing allows to discard inaccessible and costly decoding hypotheses early in the decoding. The lexicon can therefore be composed with the language model while simultaneously optimizing the resulting transducer.

Finally, the $\text{Replace}$ operation of equation~(\ref{eq:eager_replace}) is also delayed. This allows to further break the decoding graph into smaller distinct pieces: the $HCL$ transducer mapping the output of the acoustic model to words, the query language model $G_{p}$, and the entities languages models $\{G_{e_i}\, ,\forall i\in[1, n] \}$. Formally, at runtime, the dynamic decoding graph is created using the following formula

\begin{align}
(HCL)\circ \text{DynamicReplace}(G_{p}, \{G_{e_i}\, ,\forall i\in[1, n] \})\ ,
\end{align}

where the HCL transducer is computed beforehand using regular transducer compositions (i.e. $H * C * L$) and $\circ$ denotes the delayed transducer composition with composition filters.

These improvements yield real time decoding on a Raspberry Pi 3 with a small overhead compared to a \emph{static model}, and preserve decoding accuracy while reducing drastically the size of the model on disk. Additionally, this greatly reduces the training time of the LM. Finally, breaking down the LM into smaller, separate parts makes it possible to efficiently update it. It particular, performing on-device injection of new values in the LM becomes straightforward, enabling users to locally customize their SLU engine without going through the Snips web console. This feature is described in the following. 

\subsubsection{On-device personalization}

Using contextual information in ASR is a promising approach to improving the recognition results by biasing the language model towards a user-specific vocabulary~\cite{aleksic2015bringing}. A straightforward way of customizing the LM previously described is to update the list of values each entity can take. For instance, if we consider an assistant dedicated to making phone calls (``\texttt{call (Jane Doe)[contact]}''), the user's list of contacts could be added to the values of the entity ``\texttt{contact}'' in an embedded way, without this sensitive data ever leaving the device. This operation is called entity injection in the~following.

In order to perform entity injection, two modifications of the decoding graph are necessary. First, the new words and their pronunciations are added to the $HCL$ transducer. Second, the new values are added to the corresponding entity wFST $G_{e_i}$.
The pronunciations of the words already supported by the ASR are cached to avoid recomputing them on-device. Pronunciations for words absent from the $HCL$ transducer are computed via an embedded G2P. The updated $HCL$ transducer can then be fully recompiled and optimized. The procedure for adding a new value to $G_{e_i}$ varies depending on whether a union of word acceptors or an n-gram model is used. In the former case, an acceptor of the new value is created and its union with $G_{e_i}$ is computed. In the latter case, we update the n-gram counts with the new values and recompute $G_{e_i}$ using an embedded n-gram engine. The time required for the complete entity injection procedure just described ranges from a few seconds for small assistants, to a few dozen seconds for larger assistants supporting a vocabulary comprising tens of thousands of words. Breaking down the decoding graph into smaller, computationally manageable pieces, therefore allows to modify the model directly on device in order to provide a personalized user experience and increase the overall accuracy of the SLU component.

\subsubsection{Confidence scoring}
\label{sec:confidence-scoring}

An important challenge of specialized SLU systems trained on small amounts of domain-specific text data is the ability to detect out-of-vocabulary (OOV) words. Indeed, while a sufficient amount of specific training data may guarantee sampling the important words which allow to discriminate between different intents, it will in general prove unable to correctly sample filler words from general spoken language. As a consequence, a specialized ASR such as the one described in the previous sections will tend to approximate unknown words using phonetically related ones from its vocabulary, potentially harming the subsequent NLU.

\begin{figure}
  \begin{center}
    \includegraphics[width=1\columnwidth]{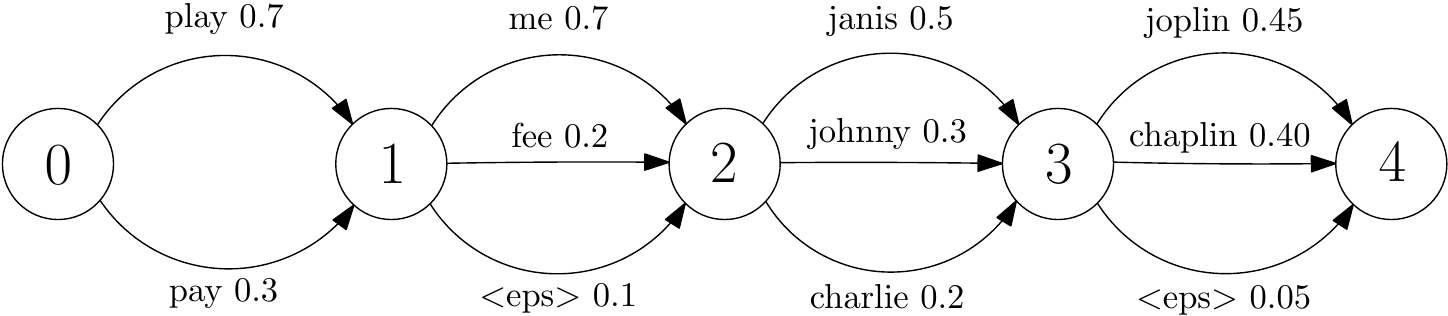}
  \end{center}{}
  \caption{Graphical representation of a confusion network. The vertices of this graph correspond to a time stamp in the audio signal, while the edges carry competing decoding hypotheses, along with their probability.\label{fig:sausage}}
\end{figure}

One way of addressing this issue is to extract a word-level confidence score from the ASR, assigning a probability for the word to be correctly decoded. Confidence scoring is a notoriously hard problem in speech recognition~\cite{jiang2005confidence,yu2011calibration}. Our approach is based on the so-called ``confusion network'' representation of the hypotheses of the ASR~\cite{mangu2000finding,xu2011minimum} (see Figure~\ref{fig:sausage}). A confusion network is a graph encoding, for each speech segment in an utterance, the competing decoding hypotheses along with their posterior probability, thus providing a richer output than the $1$-best decoding hypothesis. In particular, confusion networks in conjunction with NLU systems typically improve end-to-end performance in speech-to-meaning tasks~\cite{hakkani2006beyond,crfinslu1,tur2013semantic}. In the following, we restrict our use of confusion networks to a greedy decoder that outputs, for each speech segment, the most probable decoded word along with its~probability.

In this context, our strategy for identifying OOVs is to set a threshold on this word-level probability. Below this threshold, the word is declared misunderstood. In practice, a post-processing step is used to remove these words from the decoded sentence, replacing them with a special OOV symbol. This allows the SLU pipeline to proceed with the words the ASR has understood with sufficient probability, leaving out the filler words which are unimportant to extract the intent and the slots from the query, thus preserving the generalization properties of the SLU in the presence of unknown filler words (see section~\ref{sec:end-to-end-evaluation} for a quantitative evaluation). Finally, we may define a sentence-level confidence by simply taking the geometric mean of the word-level confidence scores.

\subsection{Natural Language Understanding}
\label{sec:nlu}

The Natural Language Understanding component of the Snips Voice Platform extracts structured data from queries written in natural language. Snips NLU -- a Python library -- can be used for training and inference, with a Rust implementation focusing solely on inference. Both have been recently open-sourced~\cite{SnipsNLU,SnipsNLURust}. 

Three tasks are successively performed. \textit{Intent Classification} consists in extracting the \textit{intent} expressed in the query (e.g. \texttt{SetTemperature} or \texttt{SwitchLightOn}). Once the intent is known, \textit{Slot Filling} aims to extract the \textit{slots}, i.e. the values of the entities present in the query. Finally, \textit{Entity Resolution} focuses on built-in entities, such as date and times, durations, temperatures, for which Snips provides an extra resolution step. It basically transforms entity values such as \texttt{"tomorrow evening"} into formatted values such as \texttt{"2018-04-19 19:00:00 +00:00"}. Snippet~\ref{lst:NLUOutput} illustrates a typical output of the NLU component.

\begin{lstlisting}[basicstyle=\ttfamily\scriptsize, float, caption={Typical parsing result for the input \texttt{"Set the temperature to 23°C in the living room"}},label={lst:NLUOutput}]
{
  "text": <@\darkgreen{"Set the temperature to 23°C in the living room"}@>,
  "intent": {
    "intentName": <@\darkgreen{"SetTemperature"}@>,
    "probability": <@\pureblue{0.95}@>
  },
  "slots": [
    {
      "entity": <@\darkgreen{"room"}@>,
      "value": <@\darkgreen{"living room"}@>,
    },
    {
      "entity": <@\darkgreen{"snips/temperature"}@>,
      "value": {
         "kind": <@\darkgreen{"Temperature"}@>,
         "unit": <@\darkgreen{"celsius"}@>,
         "value": <@\pureblue{23.0}@>
      }
    }
  ]
}
\end{lstlisting}

\begin{figure}[ht]
\vskip -0.5in
\begin{center}
\centerline{\includegraphics[width=\columnwidth]{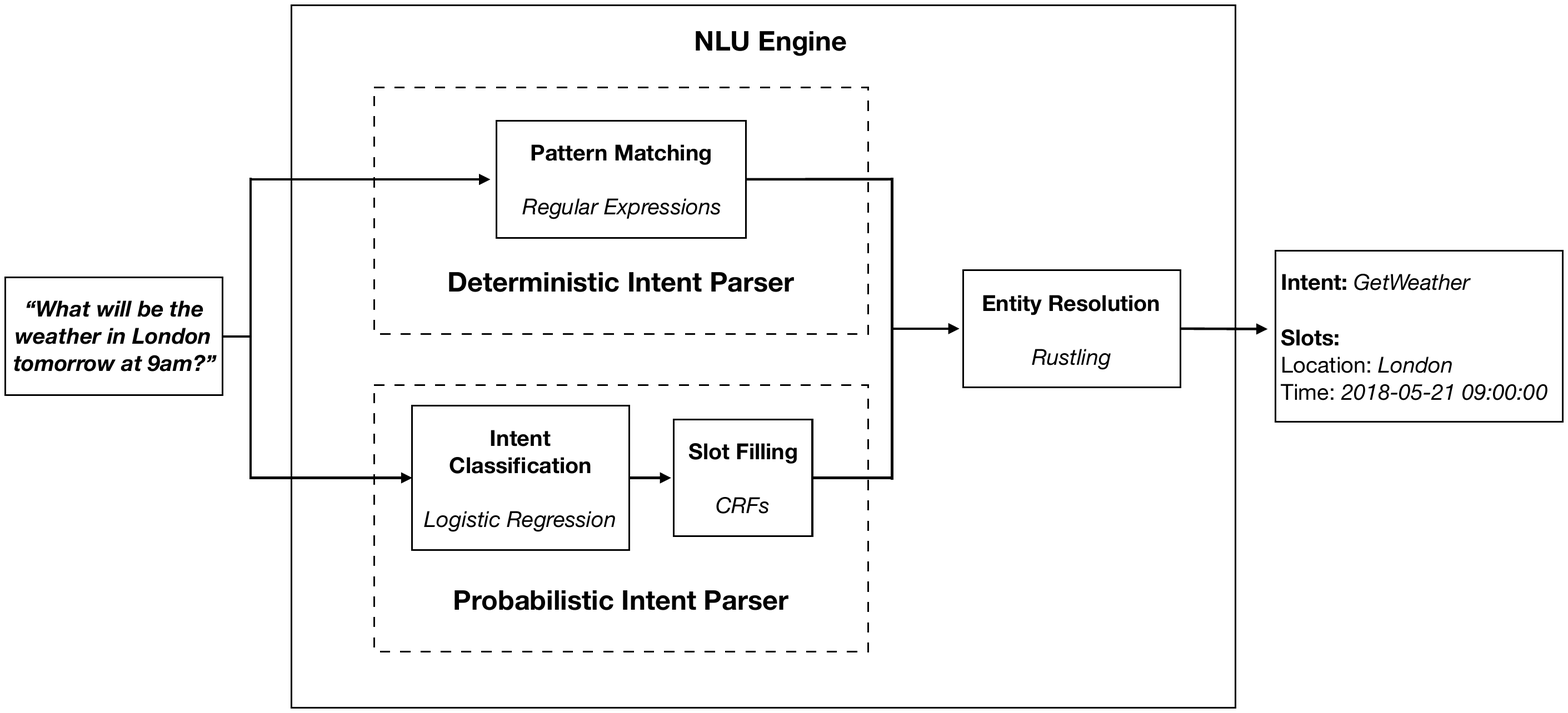}}
\vskip -0.5in
\caption{Natural Language Understanding pipeline}
\label{fig:nlu_pipeline}
\end{center}
\vskip -0.3in
\end{figure}

\subsubsection{Models}
\label{subsec:nlu_models}

The Snips NLU pipeline (\fig{nlu_pipeline}) contains a main component, the \textit{NLU Engine}, which itself is composed of several components. A first component is the \textit{Intent Parser}, which performs both intent classification and slot filling. It does not resolve entity values. The NLU Engine calls two intent parsers successively:
\begin{enumerate}
  \item a deterministic intent parser
  \item a probabilistic intent parser
\end{enumerate}
The second one is called only when nothing is extracted by the first one.

\paragraph*{Deterministic Intent Parser.}

The goal of the deterministic intent parser is to provide robustness and a predictable experience for the user as it is guaranteed to achieve a $1.0$ F1-score on the training examples.
Its implementation relies on regular expressions.
The queries contained in the training data are used to build patterns covering all combinations of entity values.
Let us consider, for instance, the training sample: 

\begin{center}
\texttt{set the [kitchen](room) lights to [blue](color)}
\end{center}

Let us assume that the set of possible values for the \texttt{room} entity are \texttt{kitchen, hall, bedroom} and those for the \texttt{color} entity are \texttt{blue, yellow, red}. A representation of the generated pattern for this sample is: 

\begin{center}
\texttt{set the (?P<room>kitchen|hall|bedroom) lights to (?P<color>blue|yellow|red)}
\end{center}

\paragraph*{Probabilistic Intent Parser.}

The probabilistic intent parser aims at extending parsing beyond training examples and recognizing variations which do not appear in the training data. It provides the generalization power that the deterministic parser lacks.
This parser runs in two cascaded steps: intent classification and slot filling. The intent classification is implemented with a logistic regression trained on the queries from every intent. The slot-filling step consists in several linear-chain Conditional Random Fields (CRFs)~\cite{crf01}, each of them being trained for a specific intent. Once the intent is extracted by the intent classifier, the corresponding slot filler is used to extract slots from the query.
The choice of CRFs for the slot-filling step results from careful considerations and experiments. They are indeed a standard approach for this task, and are known to have low generalization error~\cite{crfinslu1, crfinslu2}. Recently, more computationally demanding approaches based on deep learning models have been proposed~\cite{mesnil2013investigation,mesnil2015using}. Our experiments however showed that these approaches do not yield any significant gain in accuracy in the typical training size regime of custom voice assistants (a few hundred queries). The lightest option was therefore favored.

On top of the classical features used in slot-filling tasks such as n-grams, case, shape, etc.~\cite{tkachenko2012named}, additional features are crafted. In this kind of task, it appears that leveraging external knowledge is crucial. Hence, we apply a built-in entity extractor (see next paragraph about Entity Resolution) to build features that indicate whether or not a token in the sentence is part of a built-in entity. The value of the feature is the corresponding entity, if one is found, augmented with a BILOU coding scheme, indicating the position of the token in the matching entity value\footnote{BILOU is a standard acronym referring to the possible positions of a symbol in a sequence: Beginning, Inside, Last, Outside, Unit.}. We find empirically that the presence of such features improves the overall accuracy, thanks to the robustness of the built-in entities extractor. 

The problem of data sparsity is addressed by integrating features based on word clusters~\cite{liang2005semi}. More specifically, we use Brown clusters~\cite{brown1992class} released by the authors of~\cite{owoputi2013improved}, as well as word clusters built from word2vec embeddings~\cite{mikolov2013efficient} using k-means clustering. We find that the use of these features helps in reducing generalization error, by bringing the effective size of the vocabulary from typically 50K words down to a few hundred word clusters. 
Finally, gazetteer features are built, based on entity values provided in the training data. One gazetteer is created per entity type, and used to match tokens via a BILOU coding scheme. Table \ref{tab:crf_features} displays some examples of features used in Snips NLU.

Overfitting is avoided by dropping a fraction of the features during training. More precisely, each feature $f$ is assigned a dropout probability $p_f$. For each training example, we compute the features and then erase feature $f$ with probability $p_f$. Without this mechanism, we typically observed that the CRF learns to tag every value matching the entity gazetteer while discarding all those absent from it.

\begin{center}
\begin{table}[h]
  \hspace{-1cm}
\begin{tabular}{ l | c c c c c c c c c }
 Feature & \texttt{Will} & \texttt{it} & \texttt{rain} & \texttt{in} & \texttt{two} & \texttt{days} & \texttt{in} & \texttt{Paris} \\ \hline
$w_{-1}$ & & \texttt{Will} & \texttt{it} & \texttt{rain} & \texttt{in} & \texttt{two} & \texttt{days} & \texttt{in}\\
$brown\_cluster$ & \texttt{001110} & \texttt{011101} & \texttt{111101} & \texttt{101111} & \texttt{111111} & \texttt{111100} & \texttt{101111} & \texttt{111001}\\
$location\_entity$ & & & & & & & & \texttt{U} \\
$datetime\_builtin$ & & & & \texttt{B} & \texttt{I} & \texttt{L} &  &  \\
$number\_builtin$ & & & & & \texttt{U} & &  &  \\
$number\_builtin_{-2}$ & & & & & & & \texttt{U} &  \\
\end{tabular}
\vskip 0.2in
\caption{Examples of CRF features used in Snips NLU}
\label{tab:crf_features}
\end{table}
\vskip -0.4in
\end{center}

\paragraph*{Entity resolution.}
The last step of the NLU pipeline consists in resolving slot values (e.g. from raw strings to ISO formatted values for date and time entities). Entity values that can be resolved (e.g. dates, temperatures, numbers) correspond to the built-in entities introduced in section~\ref{sec:asr-lm-training-set}\footnote{Complete list available here: \url{https://github.com/snipsco/snips-nlu-ontology}}, and are supported natively without requiring training examples. The resolution is done with \textit{Rustling}, an in-house re-implementation of Facebook's \textit{Duckling} library~\cite{Duckling} in Rust, which we also open sourced~\cite{SnipsRustling}, with modifications to make its runtime more stable with regards to the length of the sentences parsed. 

\subsubsection{Evaluation}
\label{subsubec:nlu-eval}

Snips NLU is evaluated and compared to various NLU services on two datasets: a previously published comparison \cite{bench17}, and an in-house open dataset. The latter has been made freely accessible on GitHub to promote transparency and reproducibility\footnote{\label{note1} \url{https://github.com/snipsco/nlu-benchmark}}.

\begin{figure}[h!]
\vskip 0.2in
\begin{center}
\centerline{\includegraphics[width=\columnwidth]{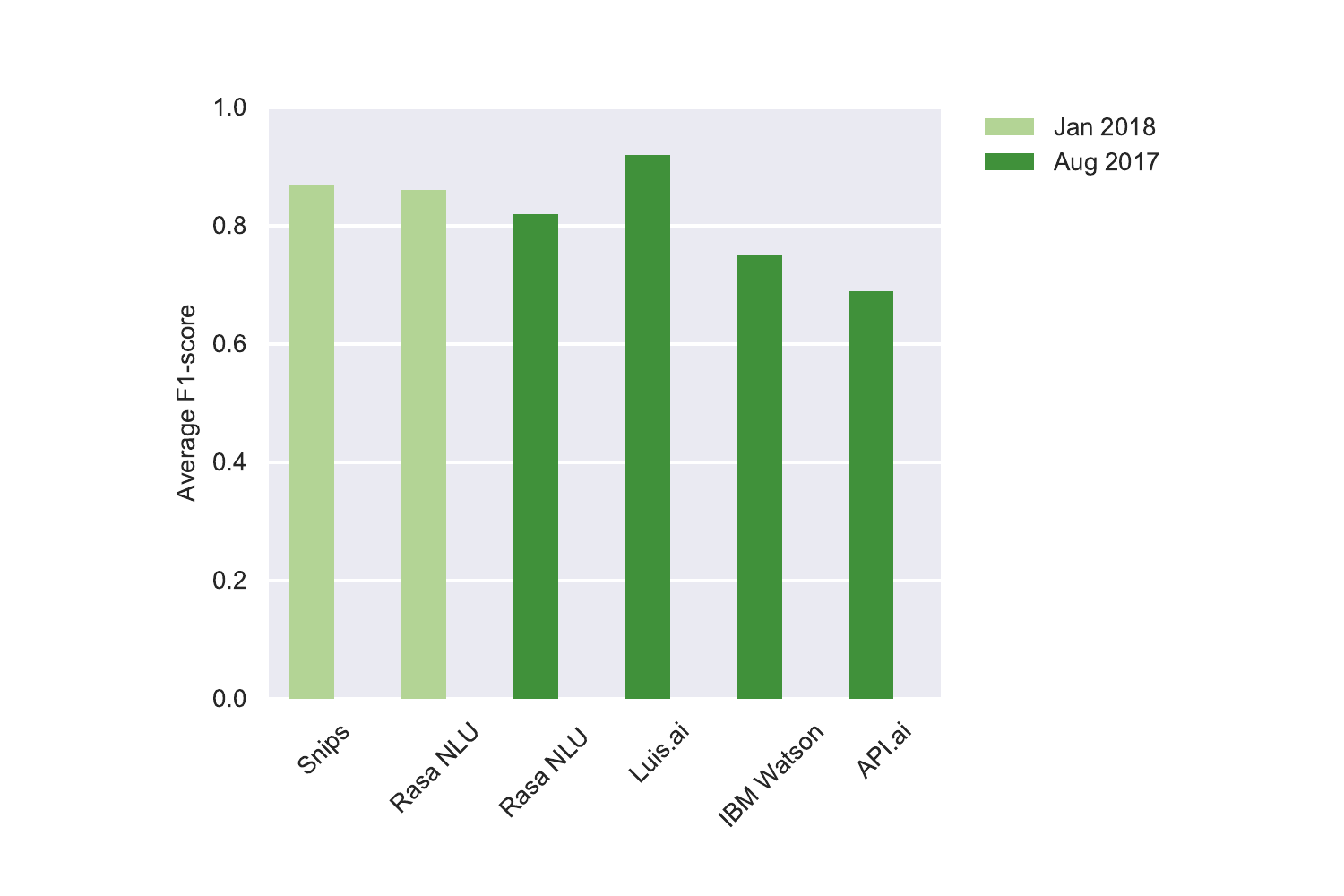}}
\vskip -0.2in
\caption{Average F1-scores, of both intent classification and slot filling, for the different NLU services over the three corpora}
\label{fig:nlu_benchmark}
\end{center}
\vskip -0.2in
\end{figure}

\paragraph*{Evaluation on Braun et al., 2017 \cite{bench17}.}
In January 2018, we evaluated Snips NLU on a previously published comparison between various NLU services~\cite{bench17}: a few of the main cloud-based solutions (Microsoft's Luis, IBM Watson, API.AI now Google's Dialogflow), and the open-source platform Rasa NLU~\cite{bocklisch2017rasa}. For the raw results and methodology, see \url{https://github.com/snipsco/nlu-benchmark}. The main metric used in this benchmark is the average F1-score of intent classification and slot filling. The data consists in three corpora. Two of the corpora were extracted from StackExchange, one from a Telegram chatbot. The exact same splits as in the original paper were used for the Ubuntu and Web Applications corpora. At the date we ran the evaluation, the train and test splits were not explicit for the Chatbot dataset (although they were added later on). In that case, we ran a 5-fold cross-validation. The results are presented in \tab{nlu_benchmark}. \fig{nlu_benchmark} presents the average results on the three corpora, corresponding to the \textit{overall} section of \tab{nlu_benchmark}. For Rasa, we considered all three possible backends (Spacy, SKLearn + MITIE, MITIE), see the abovementioned GitHub repository for more details. However, only Spacy was run on all 3 datasets, for train time reasons. For fairness, the latest version of Rasa NLU is also displayed. Results show that Snips NLU ranks second highest overall.  

\begin{center}
\begin{table}[h]
  \centering
\begin{tabular}{ | l | l | l | l | l | l |}
\hline
corpus & NLU provider & precision & recall & F1-score \\
\hline \hline
\multirow{6}{*}{chatbot} & Luis* & 0.970 & 0.918 & 0.943 \\
& IBM Watson* & 0.686 & 0.8 & 0.739 \\
& API.ai* & 0.936 & 0.532 & 0.678 \\
& Rasa* & 0.970 & 0.918 & 0.943 \\
& Rasa** & 0.933 & 0.921 & 0.927 \\
& Snips** & 0.963 & 0.899 & 0.930 \\ \hline
\multirow{6}{*}{web apps} & Luis* & 0.828 & 0.653 & 0.73 \\
& IBM Watson* & 0.828 & 0.585 & 0.686 \\
& API.ai* & 0.810 & 0.382 & 0.519 \\
& Rasa* & 0.466 & 0.724 & 0.567 \\
& Rasa** & 0.593 & 0.613 & 0.603 \\
& Snips** & 0.655 & 0.655 & 0.655 \\ \hline
\multirow{6}{*}{ask ubuntu} & Luis* & 0.885 & 0.842 & 0.863 \\
& IBM Watson* & 0.807 & 0.825 & 0.816 \\
& API.ai* & 0.815 & 0.754 & 0.783 \\
& Rasa* & 0.791 & 0.823 & 0.807 \\
& Rasa** & 0.796 & 0.768 & 0.782 \\
& Snips** & 0.812 & 0.828 & 0.820 \\ \hline
\multirow{6}{*}{overall} & Luis* & 0.945 & 0.889 & 0.916 \\
& IBM Watson* & 0.738 & 0.767 & 0.752 \\
& API.ai* & 0.871 & 0.567 & 0.687 \\
& Rasa* & 0.789 & 0.855 & 0.821 \\
& Rasa** & 0.866 & 0.856 & 0.861 \\
& Snips** & 0.896 & 0.858 & 0.877 \\ \hline
\end{tabular}
\vskip 0.2in
\caption{Precision, recall and F1-score on Braun et al. corpora. *Benchmark run in August 2017 by the authors of~\cite{bench17}. **Benchmark run in January 2018 by the authors of this paper.}
\label{tab:nlu_benchmark}
\end{table}
\end{center}

\paragraph*{Evaluation on an in-house open dataset.} In June 2017, Snips NLU was evaluated on an in-house dataset of over 16K crowdsourced queries (freely available\footnote{See footnote \ref{note1}}) distributed among 7 user intents of various complexity:
\begin{itemize}
\item \texttt{SearchCreativeWork} (e.g. Find me the I, Robot television show),
\item \texttt{GetWeather} (e.g. Is it windy in Boston, MA right now?),
\item \texttt{BookRestaurant} (e.g. I want to book a highly rated restaurant in Paris tomorrow night),
\item \texttt{PlayMusic} (e.g. Play the last track from Beyoncé off Spotify),
\item \texttt{AddToPlaylist} (e.g. Add Diamonds to my roadtrip playlist)
\item \texttt{RateBook} (e.g. Give 6 stars to Of Mice and Men)
\item \texttt{SearchScreeningEvent} (e.g. Check the showtimes for Wonder Woman in Paris)
\end{itemize}

The full ontology is available on Table \ref{tab:in-house-dataset-summary} in Appendix. In this experiment, the comparison is done separately on each intent to focus on slot filling (rather than intent classification). The main metric used in this benchmark is the average F1-score of slot filling on all slots. Three training sets of 70 and 2000 queries have been drawn from the total pool of queries to gain in statistical relevance. Validation sets consist in 100 queries per intent. Five different cloud-based providers are compared to Snips NLU (Microsoft's Luis, API.AI now Google's Dialogflow, Facebook's Wit.ai, and Amazon Alexa). For more details about the specific methodology for each provider and access to the full dataset, see \url{https://github.com/snipsco/nlu-benchmark}. Each solution is trained and evaluated on the exact same datasets. Table \ref{tab:nlu_benchmark_slot_filling} shows the precision, recall and F1-score averaged on all slots and on all intents. Results specific to each intent are available in Tables \ref{tab:slot-filling-v1} \& \ref{tab:slot-filling-v2} in Appendix. Snips NLU is as accurate or better than competing cloud-based solutions in slot filling, regardless of the training set size.

\begin{center}
\begin{table}[h]
  \centering
\begin{tabular}{| l | l | l | l | l | l | l | }
\hline
NLU provider & train size & precision & recall & F1-score \\
\hline \hline
\multirow{2}{*}{Luis}   & 70 & 0.909 & 0.537 & 0.691 \\
                        & 2000 &  0.954 & 0.917 & \textbf{0.932} \\ \hline
\multirow{2}{*}{Wit}    & 70 & 0.838 & 0.561 & 0.725 \\
                        & 2000 & 0.877 & 0.807 & 0.826 \\ \hline
\multirow{2}{*}{API.ai} & 70 & 0.770 & 0.654 & 0.704 \\
                        & 2000 & 0.905 & 0.881 & 0.884 \\ \hline
\multirow{2}{*}{Alexa}  & 70 & 0.680 & 0.495 & 0.564 \\
                        & 2000 & 0.720 & 0.592 & 0.641 \\ \hline
\multirow{2}{*}{Snips}  & 70 & 0.795 & 0.769 & \textbf{0.790} \\
                        & 2000 & 0.946 & 0.921 & 0.930 \\ \hline
\end{tabular}
\vskip 0.2in
\caption{Precision, recall and F1-score averaged on all slots and on all intents of an in-house dataset, run in June 2017.}
\label{tab:nlu_benchmark_slot_filling}
\end{table}
\end{center}

\subsubsection{Embedded performance}
Using Rust for the NLU inference pipeline allows to keep the memory footprint and the inference runtime very low. Memory usage has been optimized, with model sizes ranging from a few hundred kilobytes of RAM for common cases to a few megabytes for the most complex assistants. They are therefore fit for deployment on a Raspberry Pi or a mobile app, and more powerful servers can handle hundreds of parallel instances. Using the embedded Snips Voice platform significantly reduces the inference runtime compared to a roundtrip to a cloud service, as displayed on \tab{nlu_runtimes}.

\begin{table}[h!]
  \begin{center}
      \begin{tabular}{lr}
         \textbf{Device} & \textbf{Runtime (ms)} \\ \hline
         MacBook Pro 2.5GHz & 1.26 \\
         iPhone 6s & 2.12 \\
         Raspberry Pi 3 & 60.32 \\
         Raspberry Pi Zero & 220.02 \\
      \end{tabular}
  \vskip 0.2in
  \caption{Inference runtimes of the Snips NLU Rust pipeline, in milliseconds}
  \label{tab:nlu_runtimes}
\end{center}
\end{table}


\section{End-to-end Evaluation}
\label{sec:end-to-end-evaluation}

In this section, we evaluate the performance of the Snips Spoken Language Understanding (SLU) pipeline in an end-to-end, speech-to-meaning setting.
To this end, we consider two real-world assistants of different sizes, namely \texttt{SmartLights} and \texttt{Weather}. The \texttt{SmartLights} assistant specializes in interacting with light devices supporting different colors and levels of brightness, and positioned in various rooms. The \texttt{Weather} assistant is targeted at weather queries in general, and supports various types of formulations and places. Tables~\ref{tab:smart-lights-dataset-summary}~and~\ref{tab:weather-dataset-summary} sum up the constitution and size of the datasets corresponding to these two assistants, while tables~\ref{tab:smart-lights-entity-summary}~and~\ref{tab:weather-entity-summary} describe their entities. Note in particular the use of the built-in ``\texttt{snips/number}'' (respectively ``\texttt{snips/datetime}'') entity to define the brightness (resp. datetime) slot, which allows the assistant to generalize to values absent from the dataset.

\begin{table}[ht]
\caption{\texttt{SmartLights} dataset summary}
\label{tab:smart-lights-dataset-summary}
\hspace{-0.35cm}
\begin{tabular}{@{}lllc@{}}
Intent Name        & Slots     & Sample                                        & \#Utterances         \\ \midrule
\texttt{IncreaseBrightness} & \purple{\texttt{room}}      & Turn up the lights in the \purple{living room}         & 299                  \\
\texttt{DecreaseBrightness} & \purple{\texttt{room}}      & Turn down the lights in the \purple{kitchen}           & 300                  \\
\texttt{SwitchLightOff}            & \purple{\texttt{room}}      & Make certain no lights are on in the \purple{bathroom} & 300                  \\
\texttt{SwitchLightOn}             & \purple{\texttt{room}}      & Can you switch on my \purple{apartment} lights?        & 278                  \\
\texttt{SetLightBrightness} & \purple{\texttt{room}}      & Set the lights in the \purple{living room}             & 299                  \\
                   & \green{\texttt{brightness}} & to level \green{thirty-two}                           & \multicolumn{1}{l}{} \\
\texttt{SetLightColor}      & \purple{\texttt{room}}      & Can you change the color of the lights        & 300                  \\
                   & \red{\texttt{color}}     & to \red{red} in the \purple{large leaving room}?             & \multicolumn{1}{l}{} \\ \bottomrule
\end{tabular}
\end{table}

\begin{table}[ht]
\centering
\caption{\texttt{SmartLights} entities summary}
\label{tab:smart-lights-entity-summary}
\begin{tabular}{@{}llll@{}}
Slot Name  & Type & Range                                                 & Samples                 \\ \midrule
\texttt{room}       & custom    & 34 values                                             & kitchen, bedroom   \\
\texttt{color}      & custom    & 4 values                                              & blue, red         \\
\texttt{brightness} & built-in  & $[-10^{12}, 10^{12}]$                                     & twenty, thirty-two \\ \bottomrule
\end{tabular}
\end{table}

\begin{table}[ht]
\caption{\texttt{Weather} dataset summary}
\label{tab:weather-dataset-summary}
\hspace{-0.7cm}
\begin{tabular}{@{}lllc@{}}
Intent Name         & Slots       & Samples                                          & \#Utterances \\ \midrule
\texttt{ForecastCondition}   & \purple{\texttt{region}}      & Is it \yellow{cloudy} in \red{Germany} \green{right now}?               & 888          \\
                    & \red{\texttt{country}}     & Is \purple{South Carolina} expected to be \yellow{sunny}?          &              \\
                    & \green{\texttt{datetime}}    & Is there \yellow{snow} in \bluetwo{Paris}?                          &              \\
                    & \bluetwo{\texttt{locality}}    & Should I expect a \yellow{storm} near \orange{Mount Rushmore}?                                                   &              \\
                    & \yellow{\texttt{condition}}   &                                                  &              \\
                    & \orange{\texttt{poi}}         &                                                  &              \\
\texttt{ForecastTemperature} & \purple{\texttt{region}}      & Is it \yellow{hot} \green{this afternoon} in \red{France}?              & 880          \\
                    & \red{\texttt{country}}     & Is it \yellow{warmer} \green{tomorrow} in \purple{Texas}?                  &              \\
                    & \green{\texttt{datetime}}    & How \yellow{chilly} is it near the \orange{Ohio River}?            &              \\
                    & \bluetwo{\texttt{locality}}    & Will it be \yellow{cold} \green{tomorrow}?                        &              \\
                    & \orange{\texttt{poi}}         &                                                  &              \\
                    & \yellow{\texttt{temperature}} &                                                  &              \\
\texttt{Forecast}            & \bluetwo{\texttt{locality}}    & How's the weather \green{this morning}?                  & 877          \\
                    & \orange{\texttt{poi}}         & Forecast for \orange{Mount Rainier}                       &              \\
                    & \red{\texttt{country}}     & What's the weather like in \red{France}?               &              \\
                    & \green{\texttt{datetime}}    & Weather in \bluetwo{New York}                              &              \\
\texttt{ForecastItem}        & \purple{\texttt{region}}      & Should I wear a \yellow{raincoat} \green{in February} in \red{Canada}?  & 904          \\
                    & \yellow{\texttt{item}}        & Do I pack \yellow{warm socks} for my trip to \orange{Santorini}?   &              \\
                    & \red{\texttt{country}}     & Is a \yellow{woolen sweater} necessary in \purple{Texas} \green{in May}?  &              \\
                    & \green{\texttt{datetime}}    & Can I wear \yellow{open-toed shoes} in \bluetwo{Paris} \green{this spring}? &              \\
                    & \bluetwo{\texttt{locality}}    &                                                  &              \\
                    & \orange{\texttt{poi}}         &                                                  &              \\ \bottomrule
\end{tabular}
\end{table}

\begin{table}[ht]
\centering
\caption{\texttt{Weather} entities summary ($^\star$poi: point of interest)}
\label{tab:weather-entity-summary}
\begin{tabular}{@{}llll@{}}
Slot Name  & Type & Range                                                 & Samples                 \\ \midrule
\texttt{poi}$^\star$       & custom    & 124 values                                             & Mount Everest, Ohio River   \\
\texttt{condition}      & custom    & 28 values                                              & windy, humid          \\
\texttt{country} & custom  & 211 values                                     & Norway, France \\
\texttt{region} & custom  & 55 values                                     & California, Texas \\
\texttt{locality} & custom  & 535 values                                     & New York, Paris \\
\texttt{temperature} & custom  & 9 values                                     & hot, cold \\
\texttt{item} & custom  & 33 values                                     & umbrella, sweater \\
\texttt{datetime} & built-in & N/A & tomorrow, next month \\
\bottomrule

\end{tabular}
\end{table}

We are interested in computing end-to-end metrics quantifying the ability of the assistants to extract intent and slots from spoken utterances. We create a test set by crowdsourcing a spoken corpus corresponding to the queries of each dataset. For each sentence of the speech corpus, we apply the ASR engine followed by the NLU engine, and compare the predicted output to the ground true intent and slots in the dataset. In the following, we present our results in terms of the classical precision, recall, and F1~scores.

\subsection{Language Model Generalization Error}

To be able to understand arbitrary formulations of an intent, the SLU engine must be able to generalize to unseen queries in the same domain. To test the generalization ability of the Snips SLU components, we use 5-fold cross-validation, and successively train the LM and NLU on four fifth of the dataset, testing on the last, unseen, fifth of the data. The training procedure is identical to the one detailed in section~\ref{sec:asr-lm-language-model-building}. We note that all the values of the entities are always included in the training set. Tables~\ref{tab:smart-lights-e2e-perf}~and~\ref{tab:weather-e2e-perf} sum up the results of this experiment, highlighting in particular the modest effect of the introduction of the ASR engine compared to the accuracy of the NLU evaluated directly on the ground true query.

\begin{figure}
  \begin{center}
    \includegraphics[width=1\columnwidth]{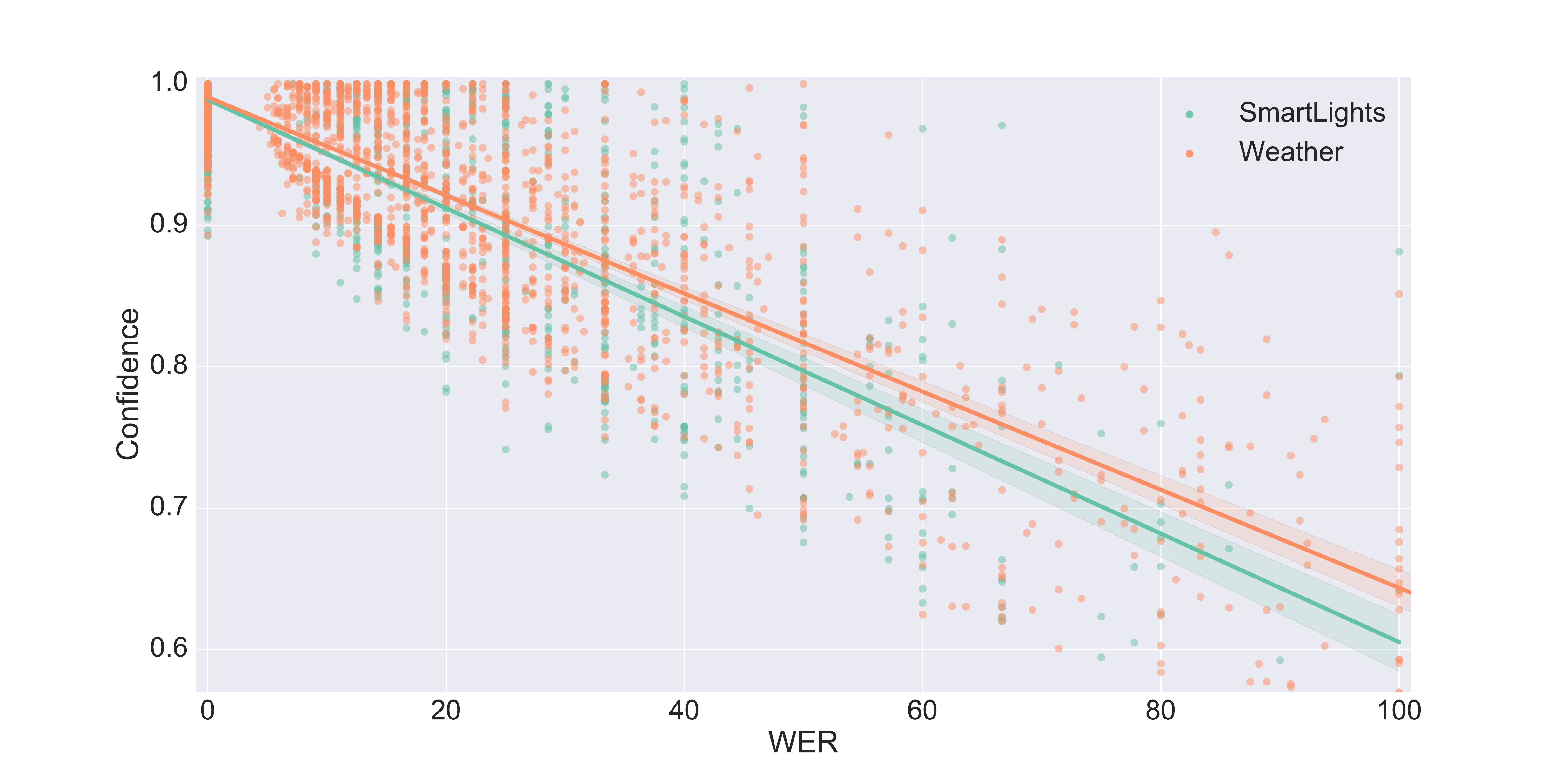}
  \end{center}
  \caption{Scatter plot of sentence-level confidence scores against word error rate. Each point represents a sentence from the test set of one of the two assistants, while the lines are obtained through a linear regression. The confidence score is correlated with the word error~rate.\label{fig:confidence_scatter}}
\end{figure}

Because unseen test queries may contain out of vocabulary words absent from the training splits, the ability of the SLU to generalize in this setting relies heavily on the identification of unknown words through the strategy detailed in section~\ref{sec:confidence-scoring}. As noted earlier and confirmed by these results, this confidence scoring strategy also allows to favor precision over recall by rejecting uncertain words that may be misinterpreted by the NLU. Figure~\ref{fig:confidence_scatter} illustrates the correlation between the sentence-level confidence score defined in section~\ref{sec:confidence-scoring} and the word error rate. While noisy, the confidence allows to detect misunderstood queries, and can be mixed with the intent classification probability output by the NLU to reject dubious queries, thus promoting precision over recall.

\begin{table}[]
\caption{End-to-end generalization performance on the \texttt{SmartLights} assistant}
\label{tab:smart-lights-e2e-perf}
\hspace{-0.9cm}
\begin{tabular}{@{}llcccclcccc@{}}
\toprule
\multicolumn{1}{c}{\multirow{2}{*}{Intent Name}} &  & \multicolumn{4}{c}{Intent Classification}  & \multicolumn{5}{c}{Slot Filling} \\ \cmidrule(lr){3-6} \cmidrule(l){7-11} 
\multicolumn{1}{c}{}                             &  & Prec         & Recall         & F1        & NLU F1  & Slot       & Prec  & Rec  & F1 & NLU F1   \\
 \cmidrule(r){1-1} \cmidrule(lr){3-6} \cmidrule(l){7-11} 
\texttt{IncreaseBrightness}                               &  & 0.92         & 0.83           & 0.87      & 0.94  & \texttt{room}       & 0.98  & 0.95 & 0.96 & 0.96 \\
\texttt{DecreaseBrightness}                               &  & 0.93         & 0.84           & 0.88      & 0.96  & \texttt{room}       & 0.97  & 0.88 & 0.93 & 0.96\\
\texttt{SwitchLightOff}                                          &  & 0.94         & 0.81           & 0.87      & 0.94 & \texttt{room}       & 0.96  & 0.95 & 0.95 & 0.98 \\
\texttt{SwitchLightOn}                                           &  & 0.88         & 0.85           & 0.86      & 0.92 & \texttt{room}       & 0.94  & 0.94 & 0.94 & 0.97 \\
\texttt{SetLightBrightness}                               &  & 0.95         & 0.84           & 0.89      & 0.97 & \texttt{room}       & 0.96  & 0.89 & 0.92 & 0.97\\
                                                 &  &              &                &           &  & \texttt{brightness} & 0.97  & 0.96 & 0.96 & 1.0\\
\texttt{SetLightColor}                                    &  & 0.90         & 0.94           & 0.92      & 0.99  & \texttt{room}       & 0.87  & 0.84 & 0.86 & 0.96 \\
                                                 &  &              &                &           &  & \texttt{color}      & 1.0   & 0.97 & 0.98 & 1.0\\ \bottomrule
\end{tabular}
\end{table}

\begin{table}[]
\caption{End-to-end generalization performance on the \texttt{Weather} assistant}
\label{tab:weather-e2e-perf}
\hspace{-1.1cm}
\begin{tabular}{@{}llccccllcccc@{}}
\toprule
\multicolumn{1}{c}{\multirow{2}{*}{Intent Name}} &  & \multicolumn{4}{c}{Intent Classification} &  & \multicolumn{5}{c}{Slot Filling}          \\ \cmidrule(lr){3-6} \cmidrule(l){8-12} 
\multicolumn{1}{c}{}                             &  & Prec    & Recall    & F1      & NLU F1    &  & Slot        & Prec & Rec  & F1   & NLU F1 \\ \cmidrule(r){1-1} \cmidrule(lr){3-6} \cmidrule(l){8-12} 
\texttt{ForecastCondition}                                &  & 0.96    & 0.89      & 0.93    & 0.99      &  & \texttt{region}      & 0.98 & 0.95 & 0.96 & 0.99   \\
                                                 &  &         &           &         &           &  & \texttt{country}     & 0.93 & 0.88 & 0.91 & 0.98   \\
                                                 &  &         &           &         &           &  & \texttt{datetime}    & 0.80 & 0.77 & 0.78 & 0.95   \\
                                                 &  &         &           &         &           &  & \texttt{locality}    & 0.92 & 0.82 & 0.87 & 0.98   \\
                                                 &  &         &           &         &           &  & \texttt{condition}   & 0.97 & 0.95 & 0.96 & 0.99   \\
                                                 &  &         &           &         &           &  & \texttt{poi}         & 0.99 & 0.86 & 0.92 & 0.98   \\
\texttt{ForecastTemperature}                              &  & 0.95    & 0.93      & 0.94    & 0.99      &  & \texttt{region}      & 0.97 & 0.94 & 0.95 & 1.0    \\
                                                 &  &         &           &         &           &  & \texttt{country}     & 0.89 & 0.88 & 0.89 & 1.0    \\
                                                 &  &         &           &         &           &  & \texttt{datetime}    & 0.78 & 0.77 & 0.78 & 0.96   \\
                                                 &  &         &           &         &           &  & \texttt{locality}    & 0.89 & 0.80 & 0.84 & 0.99   \\
                                                 &  &         &           &         &           &  & \texttt{poi}         & 0.96 & 0.84 & 0.90 & 0.99   \\
                                                 &  &         &           &         &           &  & \texttt{temperature} & 0.98 & 0.96 & 0.97 & 1.0    \\
\texttt{Forecast}                                         &  & 0.98    & 0.94      & 0.96    & 0.99      &  & \texttt{locality}    & 0.90 & 0.82 & 0.86 & 0.98   \\
                                                 &  &         &           &         &           &  & \texttt{poi}         & 0.99 & 0.93 & 0.96 & 0.98   \\
                                                 &  &         &           &         &           &  & \texttt{country}     & 0.93 & 0.90 & 0.92 & 0.98   \\
                                                 &  &         &           &         &           &  & \texttt{datetime}    & 0.80 & 0.78 & 0.79 & 0.96   \\
\texttt{ForecastItem}                                     &  & 1.0     & 0.90      & 0.95    & 1.0       &  & \texttt{region}      & 0.95 & 0.92 & 0.93 & 0.99   \\
                                                 &  &         &           &         &           &  & \texttt{item}        & 0.99 & 0.94 & 0.96 & 1.0    \\
                                                 &  &         &           &         &           &  & \texttt{country}     & 0.88 & 0.84 & 0.86 & 0.95   \\
                                                 &  &         &           &         &           &  & \texttt{datetime}    & 0.82 & 0.79 & 0.80 & 0.93   \\
                                                 &  &         &           &         &           &  & \texttt{locality}    & 0.87 & 0.80 & 0.83 & 0.98   \\
                                                 &  &         &           &         &           &  & \texttt{poi}         & 0.98 & 0.93 & 0.96 & 0.99   \\ \bottomrule
\end{tabular}
\end{table}

\subsection{Embedded Performance}

The embedded SLU components corresponding to the assistants described in the previous section are trained in under thirty seconds through the Snips web console (see section~\ref{sec:snips_ecosystem}). The resulting language models have a size of the order of the megabyte for the \texttt{SmartLights} assistant ($1.5$MB in total, with the acoustic model), and $1.5$MB for the \texttt{Weather} assistant ($1.5$MB in total). The SLU components run faster than real time on a single core on a Raspberry~Pi~3, as well as on the smaller NXP imx7D.


\section{Training models without user data}
\label{sec:datagen}

The private-by-design approach described in the previous sections requires to train high-performance machine learning models without access to users queries. This problem is especially critical for the specialized language modeling components -- Language Model and Natural Language Understanding engine -- as both need to be trained on an assistant-specific dataset. A solution is to develop a data generation pipeline. Once the scope of an assistant has been defined, a mix of crowdsourcing and semi-supervised machine learning is used to generate thousands of high-quality training examples, mimicking user data collection without compromising on privacy. The aim of this section is to describe the data generation pipeline, and to demonstrate its impact on the performance of the NLU.

\subsection{Data generation pipeline}

A first simple approach to data generation is grammar-based generation, which consists in breaking down a written query into consecutive semantic blocks and requires enumerating every possible pattern in the chosen language. While this method guarantees an exact slot and intent supervision, queries generated in this way are highly correlated: their diversity is limited to the expressive power of the used grammar and the imagination of the person having created it. Moreover, the pattern definition and enumeration can be very time consuming and requires an extensive knowledge of the given language. This approach is therefore unfit for generating queries in natural language.

On the other hand, crowdsourcing -- widely used in Natural Language Processing research~\cite{crowdsourcing} -- ensures diversity in formulation by sampling queries from a large number of demographically diverse contributors. However, the accuracy of intent and slot supervision decreases as soon as humans are in the loop. Any mislabeling of a query's intent or slots has a strong impact on the end-to-end performance of the SLU. To guarantee a fast and accurate generation of training data for the language modeling components, we complement crowdsourcing with machine-learning-based disambiguation techniques. We new detail the implementation of the data generation pipeline.

\subsubsection{Crowdsourcing}

Crowdsourcing tasks were originally submitted to Amazon Mechanical Turk\footnote{\url{https://www.mturk.com/}}, a widely used platform in non-expert annotations for natural language tasks~\cite{amt}. While a sufficient number of English-speaking contributors can be reached easily, other languages such as French, German or Japanese suffer from a comparatively smaller available crowd. Local crowdsourcing platforms therefore had to be integrated.

A text query generation task consists in generating an example of user query matching a provided set of intent and slots -- e.g. the following set: \texttt{Intent: The user wants to switch the lights on; slot: (bedroom)[room]} could result in the generated query ``\texttt{I want lights in the bedroom right now!}''. Fixing entity values reduces the task to a sentence generation and removes the need for a slot labeling step, limiting the sources of error. Diversity is enforced by both submitting this task to the widest possible crowd while limiting the number of available tasks per contributor and by selecting large sets of slot values. 

Each generated query goes through a validation process taking the form of a second crowdsourcing task, where at least two out of three new contributors must confirm its formulation, spelling, and intent. Majority voting is indeed a simple and straightforward approach for quality assessment in crowdsourcing~\cite{majorityvoting}. A custom dashboard hosted on our servers has been developed to optimize the contributor's workflow, with clear descriptions of the task. The dashboard also prevents a contributor from submitting a query that does not contain the imposed entity values, with a fuzzy matching rule allowing for inflections in all supported languages (conjugation, plural, gender, compounds, etc.).

\subsubsection{Disambiguation}

While the previous validation step allows to filter out most spelling and formulation mistakes, it does not always guarantee the correctness of the intent or the absence of spurious entities. Indeed, in a first type of errors, the intent of the query may not match the provided one -- e.g. \texttt{Switch off the lights} when the intent \texttt{SwitchLightOn} was required. In a second type of errors, spurious entities may be added by the contributor, so that they are not labeled as such -- e.g.  ``\texttt{I want lights in the \textit{guest} [bedroom](room) at \textit{60} right now!}'' when only \texttt{[bedroom](room)} was mentioned in the task specifications. An unlabeled entity has a particularly strong impact on the CRF features in the NLU component, and limits the ability of the LM to generalize. These errors in the training queries must be fixed to achieve a high accuracy of the SLU.

To do so, we perform a 3-fold cross validation of the NLU engine on this dataset. This yields predicted intents and slots for each sentence in the dataset. By repeating this procedure several times, we obtain several predictions for each sentence. We then apply majority voting on these predictions to detect missing slots and wrong intents. Slots may therefore be extended -- e.g. \texttt{(bedroom)[room]} $\rightarrow$ \texttt{(guest bedroom)[room]} in the previous example -- or added -- \texttt{(60)[intensity]} -- and ill-formed queries (with regard to spelling or intent) are filtered-out.

\subsection{Evaluation}

\begin{figure}[ht]
\vskip 0.2in
\begin{center}
\centerline{\includegraphics[width=\columnwidth]{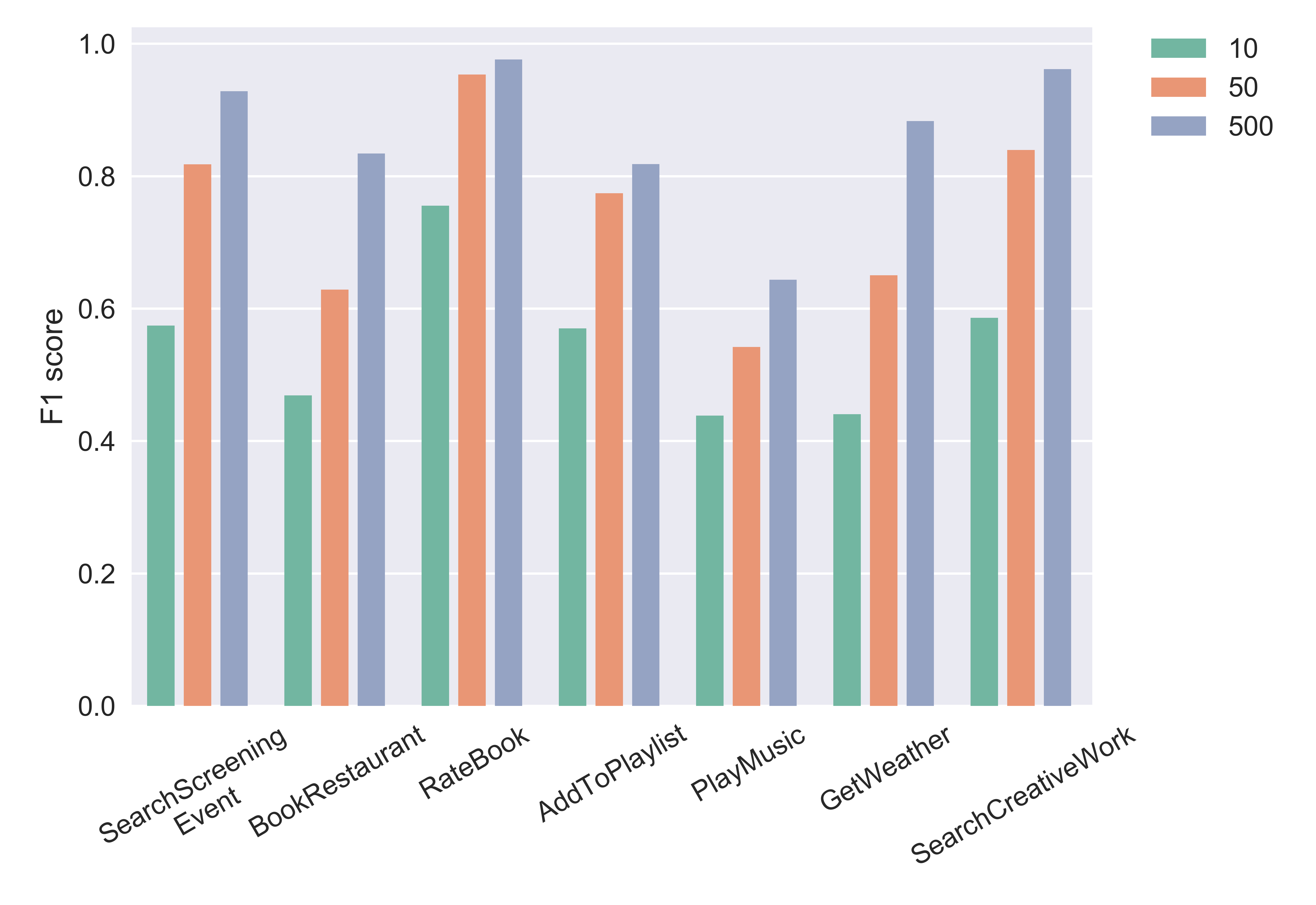}}
\caption{Average F1-score for the slot-filling task for various intents depending on the number of training queries, for 10 (green), 50 (orange), and 500 (blue) queries.}
\label{fig:datagen}
\end{center}
\vskip -0.2in
\end{figure}

We illustrate the impact of data generation on the SLU performance on the specific case of the slot-filling task in the NLU component. The same in-house open dataset of over 16K crowdsourced query presented in Section \ref{subsubec:nlu-eval} is used. Unsurprisingly, slot-filling performance drastically increases with the number of training samples. The F1-scores averaged over all slots are computed, depending on the number of training queries per intent. An NLU engine has been trained on each individual intent. Training queries are freely available on GitHub\footnote{\url{https://github.com/snipsco/nlu-benchmark}}. The data has been generated with our data generation pipeline. 

Figure~\ref{fig:datagen} shows the influence of the number of training samples on the performance of the slot-filling task of the NLU component. Compared to 10 training queries, the gain in performance with 500 queries is of 32\% absolute on average, ranging from 22\% for the \texttt{RateBook} intent (from 0.76 to 0.98) to 44\% for the \texttt{GetWeather} intent (from 0.44 to 0.88). This gain indeed strongly depends on the intent's complexity, which is mainly defined by its entities (number of entities, built-in or custom, number of entity values, etc.). While a few tens of training queries might suffice for some simple use cases (such as \texttt{RateBook}), more complicated intents with larger sets of entity values (\texttt{PlayMusic} for instance) require larger training datasets.

While it is easy to manually generate up to 50 queries, being able to come up with hundreds or thousands of diverse formulations of the same intent is nearly impossible. For private-by-design assistants that do not gather user queries, the ability to generate enough queries is key to training efficient machine learning models. Moreover, being able to generate training data allows us to validate the performance of our models before deploying them.


\section{Conclusion}

In this paper, we have described the design of the Snips Voice Platform, a Spoken Language Understanding solution that can be embedded in small devices and runs entirely offline. In compliance with the privacy-by-design principle, assistants created through the Snips Voice Platform never send user queries to the cloud and offer state-of-the-art performance. Focusing on the Automatic Speech Recognition and Natural Language Understanding engines, we have described the challenges of embedding high-performance machine learning models on small IoT devices.
On the acoustic modeling side, we have shown how small-sized neural networks can be trained that enjoy near state-of-the-art accuracy while running in real-time on small devices.
On the language modeling side, we have described how to train the language model of the ASR and the NLU in a consistent way, efficiently specializing them to a particular use case. We have also demonstrated the accuracy of the resulting SLU engine on real-world assistants.
Finally, we have shown how sufficient, high-quality training data can be obtained without compromising user privacy through a combination of crowdsourcing and machine learning.

We hope that the present paper can contribute to a larger effort towards ever more private and ubiquitous artificial intelligence. Future research directions will include private analytics, allowing to receive privacy-preserving feedback from assistant usage, and federated learning, as a complement to data generation.

\section*{Acknowledgments}

All of the work presented here has been done closely together with the engineering teams at Snips. We are grateful to the crowd of contributors who regularly work with us on the data generation pipeline. We are indebted to the community of users of the Snips Voice Platform for valuable feedback and~contributions.

\clearpage
\section*{Appendix: NLU benchmark on an in-house dataset}
\label{sec:appA}

\begin{table}[ht!]
\centering
\hspace{-0.7cm}
\begin{tabular}{@{}lllc@{}}
Intent Name         & Slots       & Samples                                          & \#Utterances \\ \midrule
\texttt{PlayMusic}   & \purple{\texttt{album, artist,}}      & I want to hear \bluetwo{I want to break free} by \purple{Queen}              & 2300          \\
                    & \purple{\texttt{track}}     & on \orange{Spotify}         &              \\
                    & \bluetwo{\texttt{playlist}}    &                                                    &              \\
                    & \yellow{\texttt{music item}}   &Play the \fuchsia{top-5} \darkgreen{soul} \yellow{songs}                                                  &              \\
                    & \orange{\texttt{service}}         &                                                   &              \\
                    & \fuchsia{\texttt{sort}}         &Put on \purple{John Lennon}'s \pureblue{1980} \yellow{album}                                                  &              \\
                    & \pureblue{\texttt{year}}         &                                                  &              \\
                    & \darkgreen{\texttt{genre}}         &                                                  &              \\
\texttt{GetWeather} & \purple{\texttt{city, country,}}      & Will it be \yellow{sunny} \orange{tomorrow} \darkgreen{near}              & 2300          \\
                    & \purple{\texttt{state}}    & \bluetwo{North Creek Forest}?           &              \\
                    & \bluetwo{\texttt{poi}}    &                          &              \\
                    & \orange{\texttt{time range}}         & How \fuchsia{chilly} is it \pureblue{here}?                                           &              \\
                    & \yellow{\texttt{condition}} &                                                           &              \\
                    & \fuchsia{\texttt{temperature}} &Should we expect \yellow{fog} in \purple{London}, \purple{UK}?                                                  &              \\
                    & \darkgreen{\texttt{spatial relation}} &                                                  &              \\
                    & \pureblue{\texttt{current location}} &                                                  &              \\
\texttt{BookRestaurant}            & \bluetwo{\texttt{sort}}    & I'd like to eat at a \pureblue{taverna} that serves                  & 2273          \\
                    & \greentwo{\texttt{party size nb}}         & \yellow{chili con carne} with a party of \greentwo{10}                       &              \\
                    & \purple{\texttt{party size descr}}    &                                                  &              \\
                    & \red{\texttt{spatial relation}}     & Make a reservation at a \bluetwo{highly rated} \pureblue{pub}  &              \\
                    & \green{\texttt{city, country}}    & for \darkgreen{tonight} in \green{Paris} \red{within walking distance}                                              &              \\
                    & \green{\texttt{state}}    & from \fuchsia{my hotel}                                                    &              \\
                    & \fuchsia{\texttt{poi}}    &                                                   &              \\
                    & \pureblue{\texttt{restaurant type}}    &Book an \blue{italian} place with a \orange{parking}                                                   &              \\
                    & \orange{\texttt{restaurant name}}    & for \purple{my grand father and I}                                                  &              \\
                    & \blue{\texttt{cuisine}}    &                                                  &              \\
                    & \yellow{\texttt{served dish}}    &                                                  &              \\
                    & \darkgreen{\texttt{time range}}    &                                                  &              \\
                    & \orange{\texttt{facility}}    &                                                  &              \\

\texttt{AddToPlaylist}        & \purple{\texttt{name, artist}}      & Add \purple{Diamonds} to \yellow{my} \red{roadtrip} playlist  & 2242          \\
                    & \yellow{\texttt{playlist owner}}        &  &              \\
                    & \red{\texttt{playlist}}     & Please add \purple{Eddy De Pretto}'s \green{album} &              \\
                    & \green{\texttt{music item}}    &  &              \\

\texttt{RateBook} & \purple{\texttt{type}}      & Rate the \yellow{current} \fuchsia{saga} \orange{three} \green{stars}             & 2256          \\
                    & \red{\texttt{name}}     &                  &              \\
                    & \green{\texttt{rating unit}}    & Rate \red{Of Mice and Men} \orange{5} \green{points} out of \bluetwo{6}          &              \\
                    & \bluetwo{\texttt{best rating}}    &                        &              \\
                    & \orange{\texttt{rating value}}         & I give the \yellow{previous} \purple{essay} a \orange{four}                                              &              \\
                    & \yellow{\texttt{select}} &                                                  &              \\
                    & \fuchsia{\texttt{series}} &                                                  &              \\
\texttt{SearchCreativeWork} & \purple{\texttt{type}}      & Find the \purple{movie} named \red{Garden State}        & 2254          \\
                    & \red{\texttt{name}}     &                  &              \\ 

\texttt{SearchScreeningEvent} & \purple{\texttt{object}}      & Which \bluetwo{movie theater} is playing            & 2259          \\
                    & \red{\texttt{type}}     &  \green{The Good Will Hunting} \yellow{nearby}?                 &              \\
                    & \green{\texttt{name}}    &           &              \\
                    & \bluetwo{\texttt{location type}}    & Show me the \purple{movie schedule} at the                        &              \\
                    & \orange{\texttt{location name}}         &\orange{Grand Rex} \fuchsia{tonight}                                               &              \\
                    & \yellow{\texttt{spatial relation}} &                                                  &              \\
                    & \fuchsia{\texttt{time range}} &                                                  &              \\ \bottomrule
\end{tabular}
\vskip 0.2in
\caption{In-house slot-filling dataset summary}
\label{tab:in-house-dataset-summary}
\end{table}

\begin{center}
\begin{table}[ht!]
\centering
\begin{tabular}{@{}llcccc@{}}
\toprule
intent & NLU provider & train size & precision &recall & F1-score \\ \midrule \midrule
\multirow{10}{*}{\texttt{SearchCreativeWork}} 
        & \multirow{2}{*}{Luis}   & 70 & 0.993 & 0.746 & 0.849 \\
        &                         & 2000 & 1.000 & 0.995 & 0.997 \\ \cmidrule(l){2-6} 
        & \multirow{2}{*}{Wit}    & 70 & 0.959 & 0.569 & 0.956 \\
        &                         & 2000 & 0.974 & 0.955 & 0.964 \\ \cmidrule(l){2-6} 
        & \multirow{2}{*}{API.ai} & 70 & 0.915 & 0.711 & 0.797 \\
        &                         & 2000 & 1.000 & 0.968 & 0.983 \\ \cmidrule(l){2-6} 
        & \multirow{2}{*}{Alexa}  & 70 & 0.492 & 0.323 & 0.383 \\
        &                         & 2000 & 0.464 & 0.375 & 0.413 \\ \cmidrule(l){2-6} 
        & \multirow{2}{*}{Snips}  & 70 & 0.864 & 0.908 & 0.885 \\
        &                         & 2000 & 0.983 & 0.976 & 0.980 \\ \midrule 
\multirow{10}{*}{\texttt{GetWeather}} 
        & \multirow{2}{*}{Luis}   & 70 & 0.781 & 0.271 & 0.405 \\
        &                         & 2000 & 0.985 & 0.902 & 0.940 \\ \cmidrule(l){2-6} 
        & \multirow{2}{*}{Wit}    & 70 & 0.790 & 0.411 & 0.540 \\
        &                         & 2000 & 0.847 & 0.874 & 0.825 \\ \cmidrule(l){2-6} 
        & \multirow{2}{*}{API.ai} & 70 & 0.666 & 0.513 & 0.530 \\
        &                         & 2000 & 0.826 & 0.751 & 0.761 \\ \cmidrule(l){2-6} 
        & \multirow{2}{*}{Alexa}  & 70 & 0.764 & 0.470 & 0.572 \\
        &                         & 2000 & 0.818 & 0.701 & 0.746 \\ \cmidrule(l){2-6} 
        & \multirow{2}{*}{Snips}  & 70 & 0.791 & 0.703 & 0.742 \\
        &                         & 2000 & 0.964 & 0.926 & 0.943 \\ \midrule
\multirow{10}{*}{\texttt{PlayMusic}} 
        & \multirow{2}{*}{Luis}   & 70  & 0.983 & 0.265 & 0.624 \\
        &                         & 2000 & 0.816 & 0.737 & 0.761 \\ \cmidrule(l){2-6} 
        & \multirow{2}{*}{Wit}    & 70 & 0.677 & 0.336 & 0.580 \\
        &                         & 2000 & 0.773 & 0.518 & 0.655 \\ \cmidrule(l){2-6} 
        & \multirow{2}{*}{API.ai} & 70 & 0.549 & 0.486 & 0.593 \\
        &                         & 2000 & 0.744 & 0.701 & 0.716  \\ \cmidrule(l){2-6} 
        & \multirow{2}{*}{Alexa}  & 70 & 0.603 & 0.384 & 0.464 \\
        &                         & 2000 & 0.690 & 0.518 & 0.546  \\ \cmidrule(l){2-6} 
        & \multirow{2}{*}{Snips}  & 70 & 0.546 & 0.482 & 0.577 \\
        &                         & 2000 & 0.876 & 0.792 & 0.823 \\ 
\bottomrule
\end{tabular}
\vskip 0.2in
\caption{Precision, recall and F1-score averaged on all slots in an in-house dataset, run in June 2017.}
\label{tab:slot-filling-v1}
\end{table}
\end{center}

\clearpage
\begin{center}
\begin{table}[ht!]
\centering
\begin{tabular}{@{}llcccc@{}}
\toprule
intent & NLU provider & train size & precision & recall& F1-score \\ \midrule  \midrule
\multirow{10}{*}{\texttt{AddToPlaylist}} 
        & \multirow{2}{*}{Luis} & 70 & 0.759 & 0.575 & 0.771\\
        &                       & 2000  & 0.971 & 0.938 & 0.953 \\ \cmidrule(l){2-6}
        & \multirow{2}{*}{Wit}    & 70    & 0.647 & 0.478 & 0.662                \\
        &          & 2000  & 0.862 & 0.761 & 0.799                \\ \cmidrule(l){2-6}
        & \multirow{2}{*}{API.ai} & 70    & 0.830 & 0.740 & 0.766                \\
        &          & 2000  & 0.943 & 0.951 & 0.947                \\ \cmidrule(l){2-6}
        & \multirow{2}{*}{Alexa}  & 70    & 0.718 & 0.664 & 0.667                \\
        &             & 2000  & 0.746 & 0.704 & 0.724                \\ \cmidrule(l){2-6}
        & \multirow{2}{*}{Snips}  & 70    & 0.787 & 0.788 & 0.785                \\
        &          & 2000  & 0.914 & 0.891 & 0.900                \\ \midrule
\multirow{10}{*}{\texttt{RateBook}} & \multirow{2}{*}{Luis}   & 70    & 0.993 & 0.843 & 0.887                \\
            &          & 2000  & 1.000 & 0.997 & 0.999                \\ \cmidrule(l){2-6}
            & \multirow{2}{*}{Wit}    & 70    & 0.987 & 0.922 & 0.933                \\
            &          & 2000  & 0.990 & 0.950 & 0.965                \\ \cmidrule(l){2-6}
            & \multirow{2}{*}{API.ai} & 70    & 0.868 & 0.830 & 0.840                \\
            &          & 2000  & 0.976 & 0.983 & 0.979                \\ \cmidrule(l){2-6}
            & \multirow{2}{*}{Alexa}  & 70    & 0.873 & 0.743 & 0.798                \\
            &          & 2000  & 0.867 & 0.733 & 0.784                \\ \cmidrule(l){2-6}
            & \multirow{2}{*}{Snips}  & 70    & 0.966 & 0.962 & 0.964                \\
            &          & 2000  & 0.997 & 0.997 & 0.997                \\ \midrule
\multirow{10}{*}{\texttt{SearchScreeningEvent}} 
            & \multirow{2}{*}{Luis}   & 70    & 0.995 & 0.721 & 0.826                \\
            &          & 2000  & 1.000 & 0.961 & 0.979                \\ \cmidrule(l){2-6} 
            & \multirow{2}{*}{Wit}    & 70    & 0.903 & 0.773 & 0.809                \\
            &          & 2000  & 0.849 & 0.849 & 0.840                \\ \cmidrule(l){2-6} 
            & \multirow{2}{*}{API.ai} & 70    & 0.859 & 0.754 & 0.800                \\
            &          & 2000  & 0.974 & 0.959 & 0.966                \\ \cmidrule(l){2-6}
            & \multirow{2}{*}{Alexa}  & 70    & 0.710 & 0.515 & 0.560                \\
            &          & 2000  & 0.695 & 0.541 & 0.585                \\ \cmidrule(l){2-6} 
            & \multirow{2}{*}{Snips}  & 70    & 0.881 & 0.840 & 0.858               \\
            &          & 2000  & 0.965 & 0.971 & 0.967                \\ \midrule
\multirow{10}{*}{\texttt{BookRestaurant}} 
            & \multirow{2}{*}{Luis}   & 70    & 0.859 & 0.336 & 0.473                \\
            &          & 2000  & 0.906 & 0.891 & 0.892                \\ \cmidrule(l){2-6}
            & \multirow{2}{*}{Wit}    & 70    & 0.901 & 0.436 & 0.597                \\
            &          & 2000  & 0.841 & 0.739 & 0.736                \\ \cmidrule(l){2-6}
            & \multirow{2}{*}{API.ai} & 70    & 0.705 & 0.548 & 0.606                \\
            &          & 2000  & 0.874 & 0.853 & 0.834                \\ \cmidrule(l){2-6} 
            & \multirow{2}{*}{Alexa}  & 70    & 0.598 & 0.364 & 0.504                \\
            &          & 2000  & 0.760 & 0.575 & 0.689                \\ \cmidrule(l){2-6} 
            & \multirow{2}{*}{Snips}  & 70    & 0.727 & 0.700 & 0.719                \\
            &          & 2000  & 0.919 & 0.891 & 0.903                \\ 
\bottomrule
\end{tabular}
\vskip 0.2in
\caption{Precision, recall and F1-score averaged on all slots in an in-house dataset, run in June 2017.}
\label{tab:slot-filling-v2}
\end{table}
\end{center}

\bibliography{paper}
\bibliographystyle{plain}

\end{document}